\pdfoutput=1

\documentclass[11pt]{article}

\usepackage[final]{acl}

\usepackage{times}
\usepackage{latexsym}
\usepackage{float}
\usepackage{placeins}
\usepackage[utf8]{inputenc} %
\usepackage[T1]{fontenc}    %
\usepackage{hyperref}       %
\usepackage{url}            %
\usepackage{booktabs}       %
\usepackage{amsfonts}       %
\usepackage{nicefrac}       %
\usepackage{microtype}      %
\usepackage{graphicx}
\usepackage{subcaption}
\usepackage{amssymb}
\usepackage{amsmath}
\usepackage{multirow}
\usepackage[table]{xcolor}
\usepackage{placeins}
\usepackage{inconsolata}
\definecolor{lightred}{RGB}{255, 200, 200}
\definecolor{lightgreen}{RGB}{220, 255, 220}
\definecolor{lightyellow}{RGB}{255, 255, 200}

\DeclareMathOperator*{\argmax}{\texttt{argmax}}

\title{Generative or Discriminative? \\ Revisiting Text Classification in the Era of Transformers}

\author{%
  Siva Rajesh Kasa, Karan Gupta, Sumegh Roychowdhury, Ashutosh Kumar, \AND Yaswanth Biruduraju,   Santhosh Kumar Kasa, Nikhil Priyatam Pattisapu, \AND Arindam Bhattacharya,  Shailendra Agarwal, Vijay Huddar
  \\
  Amazon Inc. \\
  \texttt{kasasiva,karaniis,sumegr@amazon.com} \\
}

\newcommand{\refsec}[1]{Section \ref{#1}}

\begin{document}

\maketitle

\begin{abstract}
    The comparison between discriminative and generative classifiers has intrigued researchers since \citet{Efron1975TheEO}'s seminal analysis of logistic regression versus discriminant analysis. While early theoretical work established that generative classifiers exhibit lower sample complexity but higher asymptotic error in simple linear settings, these trade-offs remain unexplored in the transformer era. We present the first comprehensive evaluation of modern generative and discriminative architectures—Auto-regressive, Masked Language Modeling, Discrete Diffusion, and Encoders for text classification. Our study reveals that the classical ``two regimes" phenomenon manifests distinctly across different architectures and training paradigms. Beyond accuracy, we analyze sample efficiency, calibration, noise robustness, and ordinality across diverse scenarios. Our findings offer practical guidance for selecting the most suitable modeling approach based on real-world constraints such as latency and data limitations.
    \footnote{Code available at: \url{https://github.com/amazon-science/Generative-vs-Discriminative-Classifiers}
    }
\end{abstract}

\section{Introduction}
\label{sec:introduction}

Text Classification (TC), a fundamental task in Natural Language Processing (NLP), encompasses various applications such as Sentiment Analysis, Topic Classification, and Emotion Detection. Since the emergence of transformer architectures, the field has been dominated by discriminative classifiers that leverage token embeddings (e.g., the \texttt{[CLS]} token in BERT \cite{devlin2019bert}). These models directly learn the conditional probability distribution $P_\theta(y|X)$, where $X$ denotes the input text and $y$ represents the ground truth label. However, as these discriminative models grow larger, they require increasingly large amounts of labeled data to achieve optimal performance, making them impractical in many real-world scenarios where labeled data is scarce or expensive to obtain \citep{pmlr-v202-zheng23f}. On the other hand, generative classifiers, which model the joint distribution $P_\theta(X,y)$, are known to work better in low-data settings, giving rise to the classical `two-regimes' phenomenon for classification \citep{Ng2001,Yogatama2017,pmlr-v202-zheng23f}. This advantage stems from their ability to learn underlying data distributions rather than just decision boundaries, allowing them to make better use of limited training examples. The inherent data efficiency of generative approaches, combined with recent advances in generative modeling such as Discrete Diffusion \citep{lou2023discrete}, motivates us to revisit the classical discriminative versus generative debate in the context of TC with Transformer-based architectures.

Prior research on generative classifiers has largely focused on non-textual tabular data, utilizing linear models such as Linear Discriminant Analysis \citep{Efron1975TheEO} and Naive Bayes \citep{Ng2001}. While \citet{Yogatama2017} extended this analysis to neural architectures using RNNs and LSTMs \citep{Hochreiter1997} for the TC task and found similar conclusions about generative advantages in low-data regimes, their study predated the transformer era. Modern NLP has seen the emergence of various successful transformer-based generative modeling paradigms, including auto-regressive (AR) models like GPT \citep{radford2018improving} that maximize likelihood directly, Discrete Diffusion models \citep{lou2023discrete} that learn through iterative denoising, and masked language models (MLM) \citep{devlin2019bert} that optimize pseudo-likelihood \cite{wang-cho-2019-bert}. These approaches offer different trade-offs in terms of computational efficiency, sample complexity, and modeling flexibility. However, a systematic comparison of these paradigms for text classification remains unexplored, particularly in the context of varying model sizes and real-world deployment considerations. Our work fills this gap by providing a practitioner-oriented study that evaluates these approaches not just on classification accuracy, but also on crucial deployment metrics including different model scales, robustness to input perturbations, reliability of output probabilities through calibration analysis, and preservation of ordinal relationships between classes. This comprehensive evaluation aims to provide concrete guidance for choosing between different generative and discriminative approaches based on specific deployment constraints and requirements.
We strategically focus on widely available public benchmark datasets for reproducibility purposes. Following \citet{li2024generative} and \citet{Yogatama2017}, our study evaluates all models trained from scratch, rather than relying on pre-trained weights, providing crucial insights for practitioners working with domain-specific data \citep{huang2019clinicalbert} or in resource-constrained environments \citep{martin-etal-2022-swahbert}. This approach helps isolate the confounding effects of the pre-training corpus \citep{razeghi2022impact} from other factors such as the modeling approach and size, which we evaluate.

Our \textbf{main contributions} include the following: \textbf{(a)} We present the first large-scale comparative study of two major classification approaches - Discriminative (Encoder) and Generative (Text Diffusion, AR, MLM) on 9 popular classification benchmark datasets, which is a first of its kind in the transformer era. Our study reveals a more nuanced interplay between model size and sample complexity than the previously known ``two regimes'' phenomenon. \textbf{(b)} We provide comprehensive analyses across multiple dimensions including \textbf{model scaling behavior, sample efficiency, and performance} in low-resource settings with \textbf{models trained from scratch}. We also introduce novel evaluation perspectives by examining \textbf{ordinal relationships between classes, output calibration and robustness to input noise}, offering insights beyond traditional classification metrics. We also evaluate these paradigms\textbf{ using pretrained models}. \textbf{(c)} Finally, we provide practical recommendations in Section~\ref{sec:conclusion} on selecting the appropriate model for deployment in various real-world scenarios, a concise summary of which is given in Table \ref{tab:model_comparison}.

\begin{figure*}[h]
  \centering
  \includegraphics[width=\textwidth]{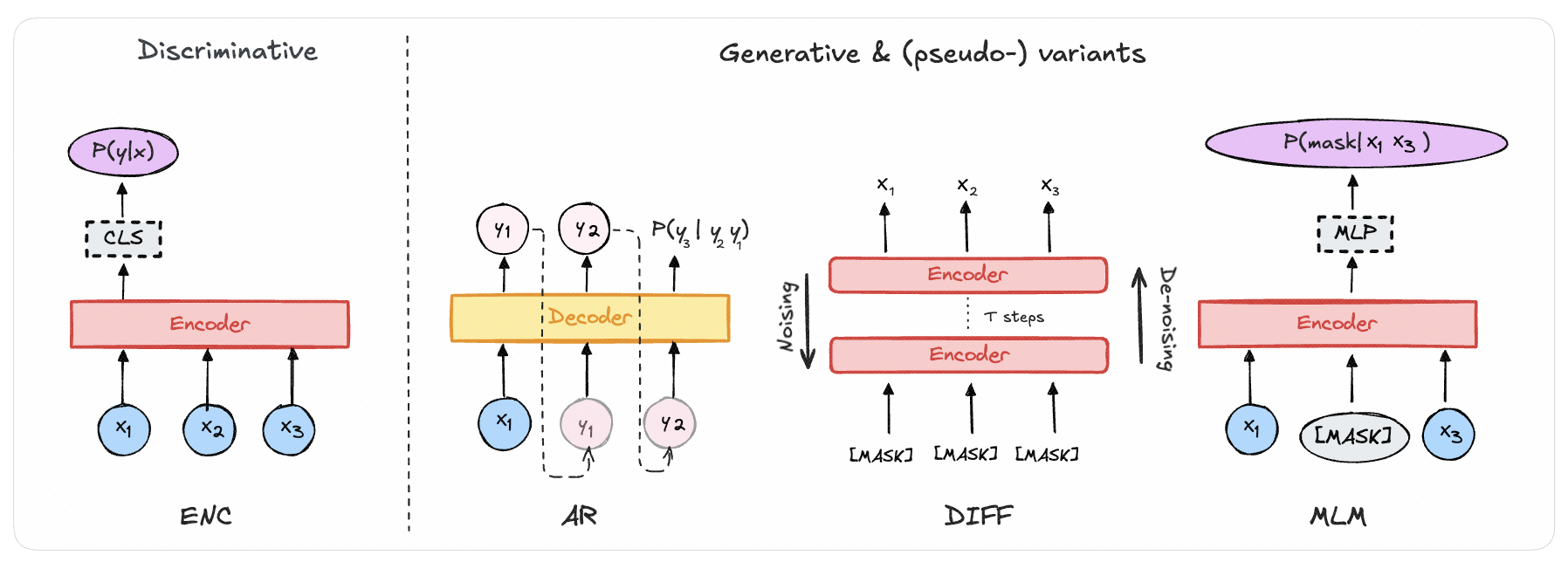}
  \caption{\small \textbf{[Best viewed in color]} Illustration of different modeling paradigms (\texttt{ENC}: Encoder-based classification, \texttt{MLM}: Masked Language Modeling, \texttt{AR}: Auto-Regressive Model, \texttt{DIFF}: Discrete Text Diffusion).}
  \label{fig:main}
\end{figure*}

\begin{table}[t]
\centering
\tiny
\begin{tabular}{p{2.5cm}p{0.5cm}p{0.5cm}p{0.6cm}p{0.5cm}p{0.5cm}}
\toprule
\textbf{Properties} & \textbf{ENC} & \textbf{AR} & \textbf{AR(p)} & \textbf{MLM} & \textbf{DIFF} \\
\midrule
Requires significant data & \cellcolor{lightyellow!50}High & \cellcolor{green!25}Low & \cellcolor{lightgreen!50}Medium & \cellcolor{lightyellow!50}High & \cellcolor{green!25}Low \\
Requires bigger model size & \cellcolor{green!25}Low & \cellcolor{lightyellow!50}High & \cellcolor{lightyellow!50}High & \cellcolor{lightyellow!50}High & \cellcolor{lightyellow!50}High \\
Sample efficiency & \cellcolor{lightyellow!50}Low & \cellcolor{green!25}High & \cellcolor{lightgreen!50}Medium & \cellcolor{lightyellow!50}Low & \cellcolor{green!25}High \\
Ordinality in scores & \cellcolor{green!25}High & \cellcolor{lightyellow!50}Low & \cellcolor{lightgreen!50}Medium & \cellcolor{green!25}High & \cellcolor{lightyellow!50}Low \\
Unimodality in Scores & \cellcolor{green!25}High & \cellcolor{lightyellow!50}Low & \cellcolor{lightgreen!50}Medium & \cellcolor{green!25}High & \cellcolor{lightyellow!50}Low \\
Well-calibrated scores & \cellcolor{green!25}High & \cellcolor{lightyellow!50}Low & \cellcolor{lightgreen!50}Medium & \cellcolor{green!25}High & N/A \\
Robustness to Noise & \cellcolor{lightgreen!50}Medium & \cellcolor{lightyellow!50}Low & \cellcolor{lightyellow!50}Low & \cellcolor{lightyellow!50}Low & \cellcolor{green!25}High \\
Inference Speed & \cellcolor{green!25}High & \cellcolor{lightgreen!50}Medium & \cellcolor{green!25}High & \cellcolor{green!25}High & \cellcolor{lightyellow!50}Low \\
\bottomrule
\end{tabular}
\caption{Comparison of different classification approaches across key properties. ENC: Encoder-based classification, AR: Auto-Regressive Model, AR(p): Pseudo-AR model, MLM: Masked Language Modeling, DIFF: Discrete Text Diffusion. Values indicate relative performance/requirements (High/Medium/Low). \textcolor{green!50}{$\blacksquare$} indicates preferred characteristics, \textcolor{yellow!70}{$\blacksquare$} indicates less favorable ones.}
\label{tab:model_comparison}
\end{table}

\section{Related Work}
\label{sec:related_work}

\textbf{Generative and Discriminative Models for Classification.}
The comparison between generative and discriminative classifiers originated with \citet{Efron1975TheEO}'s analysis of logistic regression and discriminant analysis. Building on this foundation, \citet{Ng2001} examined naive Bayes and logistic regression, establishing the fundamental trade-off between generative models' faster learning rate and discriminative models' lower asymptotic error. Their theoretical analysis heavily depends on linearity and independence assumptions. However, subsequent work by \citet{xue2008} challenged these findings through empirical studies and asymptotic analysis of statistical efficiency. \citet{Yogatama2017} provided the first empirical study of discriminative vs generative models for TC with neural architectures using LSTMs. They maximize the joint probability $P(X,y) = P(X|y) P(y)$ by concatenating the label $y$ text at the beginning of the input text $X$ and maximizing the class conditional likelihood i.e. $P(X \mid y)=\prod_{t=1}^T p\left(x_t \mid \boldsymbol{x}_{<t}, y\right)$. The final predicted label is obtained by $\hat{y} = \argmax_{y} P(X|y) P(y)$. They found that generative LSTMs have better accuracy than their discriminative counterparts at low-sample regimes. Further, they noted that the neural generative LSTMs are generally
better than baseline generative models with stronger independence assumptions (e.g. naive Bayes,
Kneser–Ney Bayes \citep{ney1994structuring,teh2006bayesian}). Next, the work by \citet{pmlr-v202-zheng23f} has extended the theoretical understanding of generative classifiers to multi-class and non-linear settings. More recent studies \citep{li2024generative,stanley2025does} have found that generative classifiers tend to avoid shortcut learning and exhibit greater robustness to distribution shifts.

While prior studies provide valuable insights, the landscape of NLP has evolved dramatically with the advent of novel transformer-based generative paradigms such as Auto-Regressive (AR) models \citep{radford2018improving} and Discrete Diffusion models \citep{lou2023discrete}. Our work extends beyond these previous comparisons by conducting the first comprehensive evaluation of modern transformer-based generative and discriminative classifiers for TC. While previous works primarily focused on classification accuracy and sample complexity, we examine multiple dimensions that are crucial for real-world deployments. For instance, \citet{Yogatama2017} initial work with neural architectures was limited to a fixed model size, leaving open questions about how the generative-discriminative trade-off varies with model capacity and computational budget—questions that have become increasingly relevant in the era of large language models. Similarly, though \citet{pmlr-v202-zheng23f} provided theoretical insights for multi-class settings, their analysis did not address practical considerations like calibration quality or preservation of ordinal relationships between classes.

\noindent \textbf{Pseudo-Generative Models.}
Recent work~\cite{sahoo2024simple} highlights a natural connection between Discrete Text Diffusion~\cite{lou2023discrete} and the Masked Language Modeling (MLM) objective in BERT~\cite{devlin2019bert}, showing that the diffusion objective can be expressed as a weighted sum of MLM losses. Using transformer encoder models, this approach achieves likelihood bounds comparable to or better than those in~\citet{lou2023discrete}. Motivated by this, we include vanilla MLM as a baseline for text classification by first appending \texttt{``The label is y"} as a suffix during training and appending \texttt{``The label is [MASK]"} as a suffix during inference. While MLM has typically served as a pretraining objective followed by fine-tuning~\cite{liu2019roberta}, there has been little systematic study of its direct use for classification. 
Although MLM does not explicitly model $P(X | y)$, it estimates $P(x_m | x_{\setminus m})$, where $x_m$ is a masked token and $x_{\setminus m}$ represents all other tokens. This approximates the pseudo-likelihood of $P(X, y)$ when modeled over the corpus \citep{wang-cho-2019-bert}. We therefore classify MLM as a pseudo-generative model.

Also, traditional generative classifiers aim to model $P(X | y)$ by prepending the label token. However, recent work~\cite{li2024generative} shows that appending the label at the end—though not strictly modeling $P(X | y)$—can yield better in-distribution performance. This setup also enables efficient inference, requiring only a single forward pass to predict the label, unlike traditional generative models that need $\#_{\text{label}}$ forward passes. These benefits motivate the inclusion of such pseudo-generative models in our benchmarks. 
Notably, these approaches involve minimal changes to standard transformer architectures—typically just altering label placement or the loss function—while preserving the core model design. This allows for fair comparisons using widely available implementations accessible to practitioners.

We also acknowledge a separate class of hybrid generative-discriminative models, where some subset of parameters are trained generatively and others discriminatively~\cite{raina2003classification,mccallum2006multi,hayashi2021}. However, we exclude them from our study, as their architectural differences hinder fair comparison with fully generative or discriminative models, placing them outside the scope of this work.

\noindent \textbf{Relation to Multi-task Learning.}
Learning $\log P(X,y)$ jointly, when factored as $\log P(X) + \log P(y|X)$ (or $\log P(y) +  \log P(X|y)$) can be viewed as a multi-task learning setup, where unsupervised learning of $\log P(X)$ ($\log P(Y)$) and supervised learning of $\log P(Y|X)$ ($\log P(X|Y)$) represent two different but related tasks. This connection is supported by empirical results showing that unsupervised pre-training helps downstream supervised tasks \citep{erhan2010does}. As demonstrated by \citet{wuunderstanding2020,hu2023revisiting}, when model capacity is sufficiently large, such multi-task learning setups tend to be more successful - the model has enough capacity to perform well on both the unsupervised and supervised objectives. However, with limited model capacity, there are inherent trade-offs between the tasks, leading to challenges in jointly optimizing for both $P(X)$ and $P(y|X)$ (or $P(y)$ and $P(X|y)$). This insight motivates us to conduct a systematic study examining the relationship between model capacity and the performance of discriminative vs generative classifiers - an analysis that has not been previously undertaken in the literature.

Refer to Appendix~\ref{app:background_related_works} for further related works review on Discrete Diffusion, Robustness to Noise, Ordinality \& Calibration.

\section{Methodology}
\label{sec:methodology}
We approach the problem of TC by leveraging two popular language modeling paradigms: \textbf{(a) Generative} - Discrete Diffusion models, Auto-regressive models (AR), and Masked Language Models (MLM)  \& \textbf{(b) Discriminative} - Encoder-based transformer models. Note that, for brevity, we use the term \textit{``generative''} from this point onward to also include the \textbf{pseudo-generative} baselines.
 Let $\mathcal{D} = \{(X_i, y_i)\}_{i=1}^N$ denote the dataset where $X_i$ is the input text and $y_i \in \mathcal{Y}$ is the corresponding label from a finite set of classes $\mathcal{Y}$. Generative models tend to learn the joint data distribution $P(X,y)$ first and then try to infer the label using the marginals, whereas Discriminative models directly learn the conditional distribution $P(y|X)$. Note that each $X_i = x_i^1 \ldots x_i^n$, where $x_i^j$ is a token from the associated vocabulary $\mathcal{V}$.

\subsection{Discriminative Model for Classification}
\label{subsec:discriminative_models}

\textbf{(1) Encoder-based classification \texttt{(ENC)}:} A Transformer encoder \cite{vaswani2017attention} $f_\theta$ encodes the input as $h_i = f_\theta(X_i)$ as a $d$-dimensional embedding, followed by a linear classifier head $W \in \mathbb{R}^{|\mathcal{Y}| \times d}$ which is the standard discriminative learning setup:
\vspace{-0.5em}
\begin{equation*}
    \hat{y}_i = \texttt{softmax}(W h_i),\mathcal{L}_{\text{enc}} = - \sum_{i=1}^{N} \log P(y_i | X_i)
\end{equation*}
where $\mathcal{L}_{\text{enc}}$ is the cross-entropy based objective for training the encoder model.

\subsection{Generative Models for Classification}
\label{subsec:generative_models}

\textbf{(2) Masked Language Modeling \texttt{(MLM)}:} During training, we first modify the input $X_i$ to : 
\begin{equation*}
X'_i = \texttt{[CLS]} \; X_i \; \texttt{[SEP]} \; \texttt{``The label is"} \;
\end{equation*} 
We then apply the standard mask i.e., on 15\% of tokens in input sequence $X'_i \oplus y_i = x_i^1 \ldots x_i^{n'} y_i$ following \citet{devlin2019bert} and predict them using unmasked bi-directional context. \citet{wang-cho-2019-bert} show that the MLM objective stochastically captures the \textit{pseudo-loglikelihood} which makes it similar to a denoising autoencoder \cite{vincent2010stacked}. Hence, we consider MLM under the generative family of models. Formally, the objective is:
\vspace{-0.5em}
\begin{equation}
    \mathcal{L}_{\text{mlm}} = - \sum_{i=1}^N \sum_{j \in \mathcal{M}_i} \log P(x_i^j | X'_i \oplus y_i ^{\setminus j})
\end{equation}
where $\mathcal{M}_i$ is the set of masked positions and $X'_i \oplus y_i ^{\setminus j}$ denotes the unmasked input with only token at position $j$ masked. At inference, we use the template: 
\begin{equation*}
    X'_i = \texttt{[CLS]} \; X_i \; \texttt{[SEP]} \; \texttt{"The label is"} \; \texttt{[MASK]} \; \texttt{.}
\end{equation*}
and predict the masked label token. The output vocabulary is restricted to the label token set $\mathcal{V}_{\mathcal{Y}}$. Since MLM returns token probabilities across the entire vocabulary for a \texttt{[MASK]} token, we extract the dimensions corresponding to the label tokens and normalize them to sum to 1, thereby obtaining the class probabilities.

\textbf{(3) Auto-regressive modeling \texttt{(AR)}:} Following \citet{radford2018improving}, we train a causal generative model to minimize the next-token prediction loss over the entire label + input sequence:
\vspace{-0.5em}
\begin{equation}
    \mathcal{L}_{\text{gpt}} = - \sum_{i=1}^N \sum_{j=1}^{L_i} \log P(x_i^j | y, x_i^1,\ldots,x_i^{j-1})
\end{equation}
where $L_i$ is the length of the $i$-th sequence. At inference time, we perform one forward pass per candidate label $y \in \mathcal{V}_\mathcal{Y}$ by prepending it to the input $X$, and compute the log-likelihood. The predicted label is then obtained as $\arg\max_{y\in\mathcal{V}_\mathcal{Y}}\log P(X \mid y)$.
In \texttt{AR$_{pseudo}$} (refer pseudo-generative models in Section~\ref{sec:related_work}) the label is appended at the end instead of the beginning and only one forward pass is required to generate the predicted label token $y$. Note that label placement is only relevant for causal generative architectures (like \texttt{AR}) with a left-to-right attention structure. For bidirectional (pseudo-)generative models like \texttt{MLM} or \texttt{DIFF}, it has no theoretical impact.

\textbf{(4) Text Diffusion \texttt{(DIFF)}:} For each input-label pair $(X_i, y_i)$, we first create a template:
\vspace{-0.5em}
\begin{equation*}
    X_i = X_i \; \texttt{[SEP]} \; \texttt{"The label is"} \; y_i \; \texttt{.}
\end{equation*}
where each template is a sequence $X_i = {x}_i^1 \ldots {x}_i^{L_i}$ with tokens ${x}_i^j \in \mathcal{V}$.

Similar to how diffusion models gradually add noise to images, our forward process gradually corrupts text by converting tokens to pure noise (here \texttt{[MASK]}). Following \citet{lou2023discrete}, we define the forward process through discrete transition matrices $Q_t$ following a continuous markov process (see eq.~\ref{eq:forward}). This process occurs at different timesteps $t \in [0,T]$, where each token position is independently corrupted, starting from the original text and progressively moving towards a completely masked sequence.
\vspace{-0.5em}
\begin{equation}
\label{eq:forward}
\frac{d p_t}{d t} = Q_t \, p_t, \quad \text{with} \quad p_0 = p_{\text{data}}
\end{equation}
The reverse process learns to reconstruct the original text by predicting what token should replace each \texttt{[MASK]} symbol. This is done by learning score ratios $s_\theta(x,t)_z = \frac{p_t(z)}{p_t(x)}$ where $x,z$ are tokens from $\mathcal{V}$ and modeling the reverse process \cite{sun2022score} as:
\begin{equation}
\label{eq:backward}
\frac{d p_{T-t}}{d t} = s_\theta(x,t)_z Q_{T-t} \, p_{T-t}
\end{equation}

\vspace{-0.5em}
\textit{Denoising Score Entropy} (DSE) is used for training the score model in a manner that ensures several desired properties for $s_\theta$ and ensures the computation is tractable:
\vspace{-0.5em}
\begin{align}
\mathcal{L}_{\text{DSE}} &= 
\mathop{\mathbb{E}}_{\substack{x_0 \sim p_0,\\ x \sim p(\cdot \mid x_0)}}
\bigg[ \sum_{z \neq x} w_{xz} \Big( s_\theta(x)_z \nonumber \\
&- \frac{p(z \mid x_0)}{p(x \mid x_0)} \log s_\theta(x)_z \Big) \bigg]
\end{align}
where $p$ is assumed to be perturbation of some base density $p_0$ and weights $w_{xz}>0$.

The ELBO (Theorem 3.6 in \citet{lou2023discrete}) provides an upper bound on the negative log-likelihood, which is what we optimize for in generative models:
\vspace{-0.5em}
\begin{equation}
    -\log p^\theta_0(x_0) \leq \mathcal{L}_{DWDSE}(x_0) + constant
\end{equation}
where $\mathcal{L}_{DWDSE}$ integrates $\mathcal{L}_{DSE}$ weighted by the forward diffusion matrix. At inference time, we mask the label token in the template $X_i$ and use the model to predict it, restricting the possible outputs to valid labels in $\mathcal{V}_{\mathcal{Y}}$. For further details, refer to \citet{lou2023discrete}.

\section{Experiments}
\label{sec:experiments}

Our experiments are designed towards addressing the following research questions:

\begin{enumerate}
\renewcommand{\labelenumi}{\textbf{Q\arabic{enumi}.}}
    \item How do different modeling approaches compare against each other when trained from scratch?
    \item How much does noise perturbation via random token substitution and token dropping affect the performance of different modeling approaches ? 
    \item How well are the different modeling approaches calibrated ? For ordinal classification, how well the predicted distributions over ordinal categories follow a unimodal shape ? 
\end{enumerate}

\subsection{Datasets}
\label{subsec:datasets}
We evaluate our models on 9 text classification benchmark datasets to ensure a comprehensive assessment across multiple domains, text lengths, and classification types - sentiment analysis, movie reviews, news categorization, and social media analysis. These are: \textbf{AG News} \cite{10.5555/2969239.2969312}, \textbf{Emotion} \cite{saravia-etal-2018-carer}, \textbf{Stanford Sentiment Treebank (SST2 \& SST5)} \cite{socher-etal-2013-recursive}, \textbf{Multiclass Sentiment Analysis}, \textbf{Twitter Financial News Sentiment}, \textbf{IMDb} \cite{maas-etal-2011-learning}, and \textbf{Hate Speech Offensive} \cite{hateoffensive}. These datasets encompass varying levels of complexity, ranging from binary text classification to fine-grained multi-class categorization, with textual inputs spanning from concise single sentences to extensive paragraph-level passages. Further details are postponed to Appendix~\ref{app:datasets}.

\subsection{Experimental Setup}
\label{subsec:experimental_setup}

We conduct an extensive benchmarking study comparing the five different modeling approaches for text classification summarised in Section~\ref{sec:methodology}: \texttt{AR}, \texttt{AR$_{pseudo}$} , \texttt{MLM}, \texttt{DIFF}, and \texttt{ENC}. These models are evaluated on 9 popular classification benchmark datasets as mentioned in Section~\ref{subsec:datasets}.

\paragraph{Checkpoint selection.}
Throughout this work, we benchmark \emph{canonical} training and model-selection pipelines for each modeling paradigm, which requires being explicit about the \emph{checkpoint selection rule} used for discriminative and generative classifiers. For discriminative encoder-based models trained with cross-entropy (negative log-likelihood, NLL), we follow the standard early-stopping protocol used throughout the classification literature (refer Appendix \ref{app:val_loss_canonical}): selecting the checkpoint that achieves the lowest validation loss. This choice is particularly appropriate in our setting because we evaluate not only hard-label metrics such as weighted-F1 but also soft-label properties including calibration and unimodality, for which log-loss is a strictly proper scoring rule and a direct measure of probabilistic quality and is widely believed to give well calibrated probabilities \citep{Gneiting2007,blasiok2023does}. Using validation log-loss for checkpoint selection aligns the selection criterion with both the training objective and the reported probability-sensitive metrics. In contrast, for autoregressive (AR) generative classifiers, there are two natural—but importantly different—choices: (i) selecting by \emph{validation language-model loss} (teacher-forced NLL on the templated sequence), which is canonical for likelihood training, or (ii) selecting by a \emph{downstream classification metric} computed from the canonical decision rule (argmax over label-conditional likelihoods; our $k$-pass argmax inference). In our AR implementation, we compute validation predictions using the canonical argmax decision rule and \emph{select the checkpoint by validation weighted-F1} (and use the same signal for early stopping), since this directly matches the classification objective of the AR classifier.

\begin{figure*}[t]
    \centering

\includegraphics[width=\textwidth]{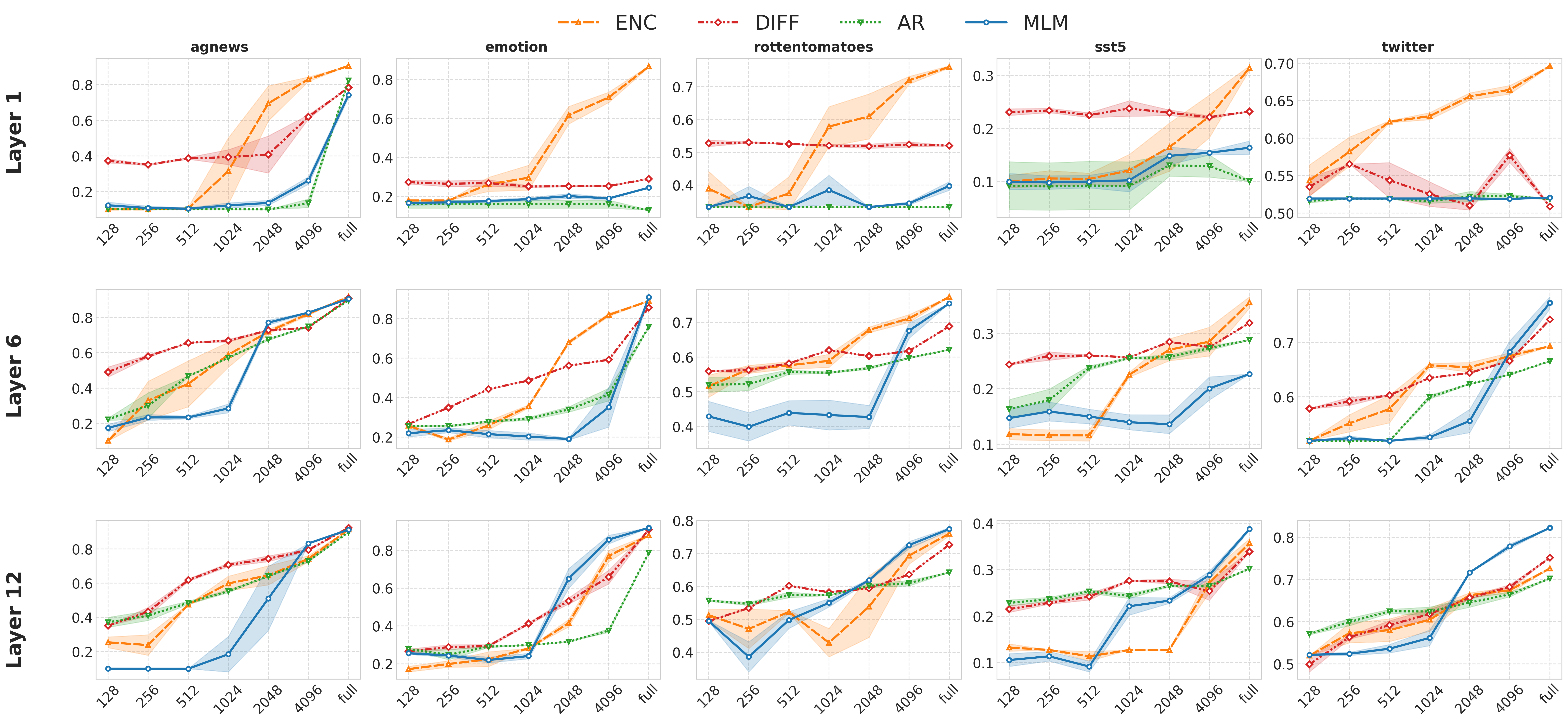}
    
    \caption{\small \textbf{[Best viewed in color]} Comparison of weighted-F1 scores of models across different configurations ($\uparrow$ is better). For rest of the datasets, refer to Figure~\ref{fig:combined_plots_all_data} in Appendix~\ref{app:main_results}. (\texttt{X-axis}: sample size, \texttt{Y-axis}: weighted-F1 score)}

    \label{fig:combined_plots}
\end{figure*}

\begin{figure*}[h]
    \centering

\includegraphics[width=\textwidth]{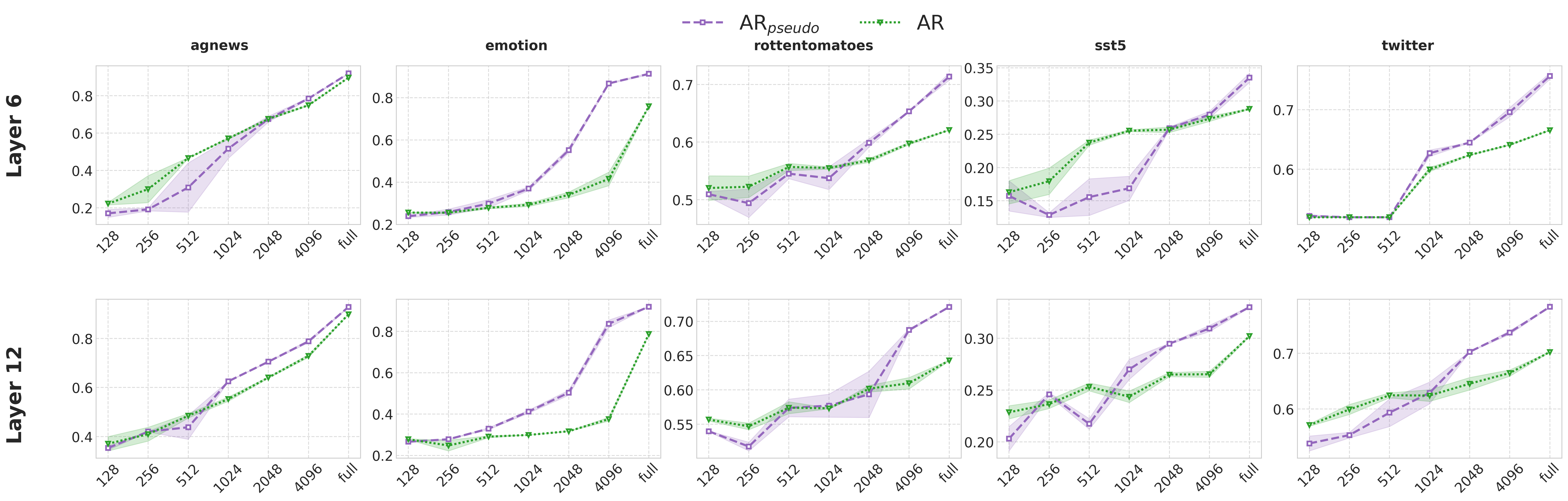}
    
    \caption{\small \textbf{[Best viewed in color]} Comparison of weighted-F1 scores between \texttt{AR$_{pseudo}$} and \texttt{AR} ($\uparrow$ is better). 1-layer results are omitted here as they are mostly trivial in low-data settings. Results for remaining datasets are provided in Figure~\ref{fig:combined_plots_gpts_all_data},  Appendix~\ref{app:main_results}. (\texttt{X-axis}: sample size, \texttt{Y-axis}: weighted-F1 score)}

    \label{fig:combined_plots_gpts}
\end{figure*}

We experiment with multiple dataset sample sizes $\in$ \{128, 256, 512, 1024, 2048, 4096, \textit{full\_data}\}.
To assess the effect of model sizes, we test 3 model size configurations using the base Transformer architecture: \textit{small} (1 layer, 1 head), \textit{medium} (6 layers, 6 heads) and \textit{large} (12 layers, 12 heads). Performance is measured using the weighted-F1 score. All experiments are repeated with 3 random seeds, running a total of $9 \times 7 \times 3 \times 3 \times 5 = 2835$ experiments and we report the average and shaded standard deviations in Figure~\ref{fig:combined_plots},~\ref{fig:combined_plots_gpts}. These experiments are designed to address \textbf{Q1}

In the second part of our evaluation, we assess each model's robustness to input perturbations. In real-world scenarios, particularly in e-commerce platforms, often encounter various text corruptions (like OCR errors in product documentation, truncated reviews, or incomplete user queries), we focus on two systematic types of synthetic noise to evaluate model robustness:
(a) \textbf{Random Token Drop} — where X\% of tokens are randomly removed from the input sentence, and
(b) \textbf{Random Token Substitution} — where X\% of tokens are replaced with random tokens from the vocabulary (excluding special tokens like \texttt{[PAD]}, \texttt{[MASK]}).
We conduct these experiments to explore \textbf{Q2} about model robustness to input perturbations.

Lastly, we assess model performance on calibration and ordinal metrics—aspects often overlooked but critical for real-world deployment. While a similar study exists for pre-trained models \citep{kasa-etal-2024-exploring}, ours focuses on models trained from scratch. For ordinal classification tasks, we verify ordinal alignment, ensuring that the predicted probability distribution reflects the natural ordering of categories (e.g., the probability of ``good'' should be closer to ``neutral'' than to ``bad'').

For ordinal evaluation, we report \textit{MSE} (Mean Squared Error), \textit{MAE} (Mean Absolute Error), and \textit{Unimodality} (UM). For calibration, we measure \textit{ECE} (Expected Calibration Error) and \textit{MCE} (Maximum Calibration Error). UM verifies that the predicted probability distribution has a single peak, thus preserving class ordering—for instance, preventing models from assigning high confidence to both extremely positive and negative sentiments simultaneously. Calibration metrics quantify the discrepancy between predicted probabilities and empirical frequencies. For detailed descriptions, see \citet{kasa-etal-2024-exploring} and \citet{wang2023calibration}. These experiments address \textbf{Q3}.
Refer to Appendix~\ref{app:implementation} for details on hyperparameters and training setup.


\section{Results}
\label{sec:results}

\begin{figure}[t]
    \centering
    \includegraphics[width=\columnwidth]{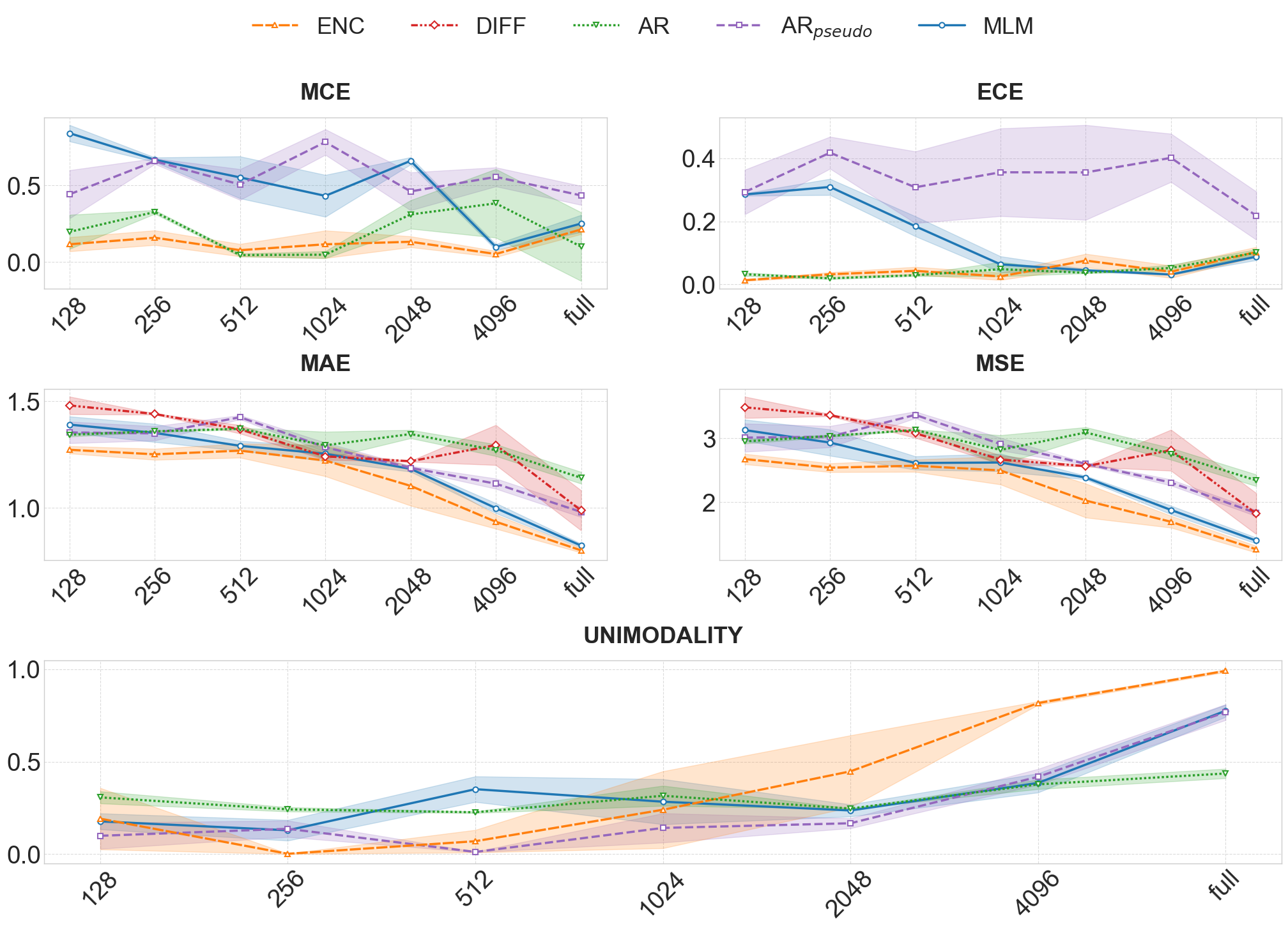}
    \caption{\small \textbf{[Best viewed in color]} Calibration and Ordinal performance of 12-layers model on SST-5. For ECE, MCE, MAE, MSE ($\downarrow$ is better) and UM ($\uparrow$ is better) (\texttt{X-axis}: sample size).}
    \label{fig:sst5_ordinal_calibration}

    \vspace{1em}

\end{figure}

\begin{figure}[t]
    \centering
    \small
    \resizebox{\columnwidth}{!}{%
    \begin{tabular}{llccccc}
    \toprule
    \textbf{Config} & Metric & 5\% & 10\% & 15\% & 20\% & 30\% \\
    \midrule
    \multirow{5}{*}{6L, 6H}
      & ENC       & \cellcolor{lightgreen!50}33.3 & 47.8 & 60.0 & 71.1 & 80.0 \\
      & AR-pseudo & 27.8 & \cellcolor{lightyellow!50}51.1 & 62.2 & 74.4 & 86.7 \\
      & AR        & 27.8 & 46.7 & \cellcolor{lightyellow!50}63.3 & \cellcolor{lightyellow!50}81.1 & \cellcolor{lightyellow!50}92.2 \\
      & MLM       & \cellcolor{lightyellow!50}32.2 & 46.7 & \cellcolor{lightyellow!50}63.3 & 72.2 & 86.7 \\
      & DIFF      & 27.8 & \cellcolor{lightgreen!50}53.3 & \cellcolor{lightgreen!50}75.6 & \cellcolor{lightgreen!50}86.7 & \cellcolor{lightgreen!50}94.4 \\
    \midrule
    \multirow{5}{*}{12L, 12H} 
      & ENC       & \cellcolor{lightyellow!50}34.4 & \cellcolor{lightyellow!50}51.1 & \cellcolor{lightyellow!50}67.8 & \cellcolor{lightyellow!50}77.8 & \cellcolor{lightyellow!50}87.8 \\
      & AR-pseudo & 33.3 & 46.7 & 61.1 & 73.3 & 86.7 \\
      & AR        & 25.6 & 37.8 & 50.0 & 67.8 & 86.7 \\
      & MLM       & 23.3 & 34.4 & 47.8 & 61.1 & 71.1 \\
      & DIFF      & \cellcolor{lightgreen!50}36.7 & \cellcolor{lightgreen!50}54.5 & \cellcolor{lightgreen!50}72.2 & \cellcolor{lightgreen!50}82.2 & \cellcolor{lightgreen!50}91.1 \\
    \bottomrule
    \end{tabular}%
    }
    \captionof{table}{\small Minimum noise\% needed for X\% weighted-F1 drop from the peak under Random Token \textbf{Dropping}. ($\uparrow$ is better)}
    \label{tab:dropping}
    
    \vspace{0.75em} %
    
    \resizebox{\columnwidth}{!}{%
    \begin{tabular}{llccccc}
    \toprule
    \textbf{Config} & Metric & 5\% & 10\% & 15\% & 20\% & 30\% \\
    \midrule
    \multirow{5}{*}{6L, 6H} 
      & ENC       & \cellcolor{lightgreen!50}26.7 & \cellcolor{lightgreen!50}37.8 & \cellcolor{lightgreen!50}51.1 & \cellcolor{lightgreen!50}58.9 & \cellcolor{lightgreen!50}76.7 \\
      & AR-pseudo & 15.6 & 21.1 & 27.8 & 32.2 & 50.0 \\
      & AR        & 21.1 & 30.0 & 38.9 & 47.8 & 62.2 \\
      & MLM       & 20.0 & 32.2 & \cellcolor{lightyellow!50}44.4 & \cellcolor{lightyellow!50}51.1 & 63.3 \\
      & DIFF      & \cellcolor{lightyellow!50}22.2 & \cellcolor{lightyellow!50}34.4 & 41.1 & 50.0 & \cellcolor{lightyellow!50}75.6 \\
    \midrule
    \multirow{5}{*}{12L, 12H} 
      & ENC       & \cellcolor{lightgreen!50}22.2 & \cellcolor{lightyellow!50}32.2 & \cellcolor{lightyellow!50}42.2 & \cellcolor{lightyellow!50}47.8 & 61.1 \\
      & AR-pseudo & 13.3 & 22.2 & 27.8 & 34.4 & 51.1 \\
      & AR        & 20.0 & 28.9 & 38.9 & \cellcolor{lightgreen!50}52.2 & \cellcolor{lightyellow!50}67.8 \\
      & MLM       & \cellcolor{lightyellow!50}21.1 & 31.1 & 38.9 & 44.4 & 55.6 \\
      & DIFF      & 16.7 & \cellcolor{lightgreen!50}35.6 & \cellcolor{lightgreen!50}44.4 & \cellcolor{lightgreen!50}52.2 & \cellcolor{lightgreen!50}73.3 \\
    \bottomrule
    \end{tabular}%
    }
    \captionof{table}{\small Minimum noise\% needed for X\% weighted-F1 drop from the peak under Random Token \textbf{Substitution}. ($\uparrow$ is better)}
    \label{tab:substitution}
\end{figure}

\vspace{-0.5em}
We analyze the results from all the experiments and provide valuable insights \& recommendations for model selection.

\noindent \textbf{Q1:} For \textbf{1-layer, 1-head} models (Figure~\ref{fig:combined_plots}), all approaches show near-random performance in low-data regimes. However, as training data increases, only \texttt{ENC} (orange line) continues to improve, ultimately outperforming others in high-data settings. This suggests that \textbf{for small models - often necessary due to real-world latency constraints - \texttt{ENC} is the most effective approach.} The classical `two regimes' phenomenon does not manifest when the model size is small.

The pattern shifts dramatically for larger architectures.%
Under the \textbf{12-layer, 12-head} configuration, both generative models—\texttt{AR} and \texttt{DIFF}—outperform \texttt{ENC} in low-data settings, with this advantage diminishing as data increases. This aligns with previous findings \citep{Ng2001,Yogatama2017,rezaee2021discriminative} about generative models' advantages in data-limited scenarios. 
Surprisingly, for large models, the pseudo-generative \texttt{MLM} (\textit{blue} line) consistently outperforms all methods across our 9 benchmark datasets in high-data settings, \textbf{challenging the conventional wisdom about discriminative dominance in high-sample regime.} This aligns with \citet{erhan2010does}'s finding that pseudo-generative models implicitly perform unsupervised pre-training alongside supervised learning, creating an effective multi-task setup (Section~\ref{sec:related_work}). Their work shows that this unsupervised phase acts as a data-dependent regularizer, guiding optimization toward better-generalizing minima. For \textit{large} models, direct fine-tuning without this implicit pre-training often leads to suboptimal convergence, explaining \texttt{ENC}'s underperformance relative to \texttt{MLM}. Thus, \textbf{for scenarios without model size constraints, generative models emerge as the optimal choice for low-data settings} such as for low-resource languages and continual learning applications requiring frequent updates with limited samples, while \textbf{pseudo-generative \texttt{MLM} is superior when abundant labeled data is available.}

Another noteworthy observation is that under the \textbf{6-layer, 6-head} configuration, in low-data settings, \texttt{DIFF} emerges as the best performing model across all datasets, clearly outperform even it's generative counterpart \texttt{AR}. As the training data size increases, we see that the discriminative \texttt{ENC} outperforming \texttt{DIFF}. Thus, \textbf{in medium scale architectures, between the generative \texttt{DIFF} and the discriminative \texttt{ENC}, the classical `two regimes' still holds.} 

Figure~\ref{fig:combined_plots_gpts} shows that \texttt{AR$_{pseudo}$} generally underperforms \texttt{AR} and also displays \textbf{higher variance} in low-data settings—the recommended use case—while the opposite holds in high-data scenarios. This reveals a new insight beyond \citet{li2024generative}, who only evaluated full-data settings where \texttt{AR$_{pseudo}$} performed better in-distribution. As noted in Section~\ref{sec:methodology}, \texttt{AR} requires $|label|$-times forward passes per prediction, unlike the single pass needed for \texttt{AR$_{pseudo}$}; however, this can be mitigated via batching or parallel processing, reducing inference time differences at the cost of higher computation. We also investigate the claim that ``larger models can sometimes deteriorate performance'' \citep{nakkiran2019deepdoubledescentbigger} in the Appendix~\ref{sec:revisiting_bias_variance}. While we observe the classical bias-variance trade-off in small-data regimes, performance generally improves with model size in full-data settings especially for AR.

\paragraph{Why the best-loss and best-F1 checkpoints can differ (especially in low-data).} 
Here we clarify the log-loss minimizing checkpoint selection further over and above the reasons of proper scoring rule and calibration sensitive evaluation discussed in \S \ref{subsec:experimental_setup}. 
In low-data regimes, the checkpoint minimizing validation log-loss need not coincide with the checkpoint maximizing weighted-F1. This is expected because log-loss is sensitive to the full predictive distribution and penalizes \emph{a few extremely confident errors} heavily as log-loss is unbounded \citep{quinonero2005evaluating, pmlr-v70-guo17a}, whereas F1 depends only on the argmax decision boundary; thus a model can improve F1 while simultaneously worsening log-loss as it becomes increasingly confident on a shrinking set of remaining errors. This divergence is particularly pronounced when models memorize small training sets and produce overconfident predictions. Such highly confident memorization behavior is also connected to privacy risk \citep{DBLP:conf/csfw/YeomGFJ18}: membership-inference attacks explicitly exploit differences between training and non-training points (often reflected in loss/confidence gaps), and stronger overfitting increases membership-inference vulnerability. Therefore, while selecting checkpoints by validation weighted-F1 can yield higher F1 , this comes with the potential cost of poorer probabilistic reliability and increased privacy vulnerability. Hence, for discriminative classifiers, we stick with the canonical log-loss minimizing checkpoint selection approach.

\noindent \textbf{Q2:} We evaluate the robustness of all approaches under both 6-layer and 12-layer configurations across two noise schedules in full-data settings. We exclude 1-layer models from this analysis since their performance is mostly trivial (except for \texttt{ENC}), making robustness comparisons uninformative. In Tables ~\ref{tab:dropping} and ~\ref{tab:substitution}, we report the minimum noise level required to degrade a model's performance by a certain threshold $X\%=\{5\%,10\%,\dots\}$ relative to its peak, averaged across all datasets, as a measure of \textbf{robustness boundary}. Our analysis reveals that all models exhibit lower robustness to substitution noise compared to dropping. This can be explained by the inequality: $P(\text{garbage}_t|X_{1...t-1})<P(x_{t+1}|X_{1...t-1})$—the model is more likely to assign lower probability to a corrupted token than to a skip token at $t+1$-th position (assuming $t$-th token was dropped), which may still be contextually relevant given $X_{1...t-1}$.

The generative \texttt{DIFF} demonstrates superior robustness to both token dropping and substitution (except in 6-layers where \texttt{ENC} is slightly better), likely because its training paradigm involves recovering true tokens from noise/masked inputs. The discriminative \texttt{ENC} maintains consistent robustness under both noise types, while generative \texttt{AR} shows the high sensitivity to noise. Combining these findings with \textbf{Q1}'s results reveals that generative \texttt{AR} models face dual challenges compared to \texttt{ENC} in full data settings: they underperform in terms of both weighted-f1 and robustness. This contrasts with \citet{li2024generative}'s findings that discriminative \texttt{ENC} models rely on shortcuts and show less robustness compared to generative \texttt{AR}. However, their analysis focuses on shortcut learning and distribution shifts rather than input perturbation noise across varying model sizes. Notably, while the pseudo-generative \texttt{MLM} and \texttt{AR$_{pseudo}$} demonstrate superior performance in larger models at full data settings, they exhibit lower robustness compared to similarly performing \texttt{ENC} models. Moreover the relative drop in robustness in moving from dropping to substitution noise is more severe in \texttt{AR$_{pseudo}$} compared to \texttt{AR}. This is likely because \texttt{AR$_{pseudo}$} conditions on corrupted inputs, so it’s directly affected by garbage tokens polluting the predictive context. However, for \texttt{AR} clean label conditions the model, and the noisy input is scored globally — giving the model more flexibility to discount garbage.

\noindent \textbf{Q3:} Figure~\ref{fig:sst5_ordinal_calibration} presents ordinal and calibration results for SST-5, selected for its balanced distribution, inherent class ranking (e.g., very positive to very negative), and highest number of classes. Results for other datasets are in Appendix~\ref{app:ordinal_calibration}. \texttt{DIFF} does not support calibration metrics like ECE, MCE, and UM, as its masking/absorbing noise process produces only binary outputs rather than soft probabilities. While a uniform noise schedule can yield probabilities over $\mathcal{V}$, it performed slightly worse, so we used the absorbing schedule in our study.

From the ECE and MCE plots, we observe that \texttt{ENC} outputs remain well-calibrated across all sample sizes, while \texttt{MLM} reaches similar calibration only in high-data regimes. 
We also see that \texttt{MLM} and \texttt{ENC} achieve UM in over 80\% of the samples, aligning with findings from \citet{kasa-etal-2024-exploring}. Their MAE and MSE values are also low, indicating strong ordinality in high-data settings. This completes the picture for \textit{large} models under high-data, where \texttt{MLM} not only outperforms others in weighted-F1 but is also well-calibrated and ordinal, making it a strong candidate for real-world deployment. 
However, under low-data conditions, 12-layers \texttt{AR} outperforms \texttt{AR$_{pseudo}$} in 7 out of 9 datasets on calibration metrics. It also surpasses \texttt{DIFF} in ordinal performance, thus making it the more reliable choice among generative models in low-data scenario. Also, even though generative approaches like \texttt{DIFF} were recommended earlier in \textbf{Q1} based on weighted-F1 for 6-layers case (in Figure~\ref{fig:combined_plots}) deploying them in production could be risky when calibrated or ordinal probabilities are required, especially for imbalanced datasets like \textit{twitter} and \textit{hatespeech} (see Appendix~\ref{app:ordinal_calibration}). These metrics are particularly important when downstream models consume output probability scores as features which is often the case in multi-stage ranking systems.

Lastly, Figure~\ref{fig:calibration_ordinal_across_layers} in Appendix~\ref{app:ordinal_calibration} reveals an interesting trend: as \textit{model size} increases, calibration metrics either remain flat or worsen. This suggests that larger models or improved classification accuracy do not necessarily lead to better calibration, aligning with the findings of \citet{pmlr-v70-guo17a} where they show similar behaviour using ResNets \citep{he2016deep}. However, for ordinal metrics, we observe substantial improvements when moving from 1-layer to 6-layer models, with performance plateauing at 12 layers. A similar trend was reported in \citet{kasa-etal-2024-exploring} for pre-trained models.

\subsection{Impact of Initialization with Pretrained models}
To investigate the impact of pretraining, we conducted additional experiments using BERT-base (for ENC) and GPT-2 base (for AR), both featuring comparable architectures with 12 layers and 12 attention heads. These models were fine-tuned on our benchmark datasets across various sample sizes, maintaining consistency with our previous experimental protocol. Figure \ref{fig:pretrained_plot} has the detailed results.

These experiments reveal important insights that contrast with our findings from models trained from scratch. When using pretrained weights, we observe that the classical ``two regimes" phenomenon no longer holds. Instead, the discriminative ENC model consistently outperforms the generative AR approach across all data regimes in most datasets. This aligns with recent findings from \citet{pmlr-v202-zheng23f} in the vision domain, where pretraining was shown to eliminate the two-regime effect. This behavior can be theoretically explained by viewing pretraining as providing models with ``asymptotically large" amounts of data, effectively reducing the traditional advantage that generative models hold in low-data settings since both architectures begin with rich, generalized representations. However, these results should be interpreted with several caveats in mind: pretrained models often employ mixed training objectives (e.g., BERT uses both MLM and Next Sentence Prediction (NSP)), rely on different pretraining datasets with varying cutoff dates, and have distinct architectural designs. Additionally, our pretrained analysis was limited to comparing AR and ENC models due to the current unavailability of pretrained diffusion models.

\section{Conclusion}
\label{sec:conclusion}
Our study offers practical modeling recommendations across deployment scenarios. For latency-sensitive applications, \texttt{ENC} is ideal—especially in the 1-layer setting—due to its efficiency, robustness to noise, and well-calibrated, ordinal outputs. For offline settings with sufficient data, the 12-layer \texttt{MLM} performs best across F1, calibration, and ordinal metrics, though caution is needed with noisy inputs due to its lower robustness to token dropping. In low-resource scenarios, both \texttt{AR} and \texttt{DIFF} are strong options, with \texttt{DIFF} favored for its noise resilience and performance at 6-layers. However, if calibrated probability outputs are essential, such as in ranking pipelines, \texttt{AR} is the preferred choice.

\section{Limitations}
\label{sec:limitations}

While we conducted a thorough examination of generative and discriminative classifiers under standard i.i.d. assumptions, our findings may not generalize to scenarios involving distribution shifts, such as co-variate shift \citep{bickel2009discriminative} or concept shift \citep{roychowdhury2024tackling}. Our analysis was limited to traditional fine-tuning approaches, excluding emerging paradigms such as few-shot prompt-based in-context learning \citep{sun2023text,gupta2023robust} and parameter-efficient techniques like LoRA \citep{hu2022lora}, which may uncover newer insights. Furthermore, our study focused exclusively on pure text classification, leaving the exploration of multi-modal scenarios involving tabular data \citep{pattisapu2025leveraging}, images \cite{lu2019vilbert}, audio \cite{kushwaha2023multimodal}, and other modalities for future work.

\bibliography{references}

@article{Yogatama2017,
  author  = {Dani Yogatama and Chris Dyer and Wang Ling and Phil Blunsom},
  title   = {Generative and Discriminative Text Classification with Recurrent Neural Networks},
  journal = {arXiv preprint},
  year    = {2017}
}

@inproceedings{Ng2001,
  author    = {Andrew Y. Ng and Michael I. Jordan},
  title     = {On discriminative vs. generative classifiers: A comparison of logistic regression and naive Bayes},
  booktitle = {Advances in Neural Information Processing Systems},
  year      = {2001}
}

@article{Hochreiter1997,
  author  = {Sepp Hochreiter and J{\"u}rgen Schmidhuber},
  title   = {Long short-term memory},
  journal = {Neural Computation},
  volume  = {9},
  number  = {8},
  pages   = {1735--1780},
  year    = {1997}
}

@incollection{prechelt2012earlystop,
  author    = {Lutz Prechelt},
  title     = {Early Stopping --- But When?},
  booktitle = {Neural Networks: Tricks of the Trade},
  editor    = {Gr{\'e}goire Montavon and Genevi{\`e}ve B. Orr and Klaus-Robert M{\"u}ller},
  series    = {Lecture Notes in Computer Science},
  volume    = {7700},
  pages     = {53--67},
  publisher = {Springer, Berlin, Heidelberg},
  year      = {2012},
  doi       = {10.1007/978-3-642-35289-8_5}
}

@techreport{prechelt1997earlystop,
  author      = {Lutz Prechelt},
  title       = {Early Stopping --- But When?},
  institution = {Fakult{\"a}t f{\"u}r Informatik, Universit{\"a}t Karlsruhe},
  year        = {1997},
  url         = {https://page.mi.fu-berlin.de/prechelt/Biblio/stop_tricks1997.pdf}
}

@article{DBLP:journals/corr/abs-1206-5533,
  author       = {Yoshua Bengio},
  title        = {Practical recommendations for gradient-based training of deep architectures},
  journal      = {CoRR},
  volume       = {abs/1206.5533},
  year         = {2012},
  url          = {http://arxiv.org/abs/1206.5533},
  eprinttype   = {arXiv},
  eprint       = {1206.5533},
  bibsource    = {dblp computer science bibliography, https://dblp.org}
}

@inproceedings{PhamQWLRTYN20,
  title     = {Problems and Opportunities in Training Deep Learning Software Systems: An Analysis of Variance},
  author    = {Hung Viet Pham and Shangshu Qian and Jiannan Wang and Thibaud Lutellier and Jonathan Rosenthal and Lin Tan and Yaoliang Yu and Nachiappan Nagappan},
  booktitle = {35th IEEE/ACM International Conference on Automated Software Engineering, ASE 2020, Melbourne, Australia, September 21--25, 2020},
  pages     = {771--783},
  year      = {2020},
  publisher = {IEEE},
  isbn      = {978-1-4503-6768-4},
  doi       = {10.1145/3324884.3416545},
  url       = {https://www.cs.purdue.edu/homes/lintan/publications/variance-ase20.pdf}
}

@article{lu2021clam,
  author  = {Lu, Ming Y. and Williamson, Drew F. K. and Chen, Tiffany Y. and Chen, Richard J. and Barbieri, Matteo and Mahmood, Faisal},
  title   = {Data-efficient and weakly supervised computational pathology on whole-slide images},
  journal = {Nature Biomedical Engineering},
  volume  = {5},
  pages   = {555--570},
  year    = {2021},
  doi     = {10.1038/s41551-020-00682-w},
  url     = {https://www.nature.com/articles/s41551-020-00682-w}
}

@InProceedings{pmlr-v162-mindermann22a,
  title     = {Prioritized Training on Points that are Learnable, Worth Learning, and not yet Learnt},
  author    = {Mindermann, S{\"o}ren and Brauner, Jan M and Razzak, Muhammed T and Sharma, Mrinank and Kirsch, Andreas and Xu, Winnie and H{\"o}ltgen, Benedikt and Gomez, Aidan N and Morisot, Adrien and Farquhar, Sebastian and Gal, Yarin},
  booktitle = {Proceedings of the 39th International Conference on Machine Learning},
  pages     = {15630--15649},
  year      = {2022},
  editor    = {Chaudhuri, Kamalika and Jegelka, Stefanie and Song, Le and Szepesvari, Csaba and Niu, Gang and Sabato, Sivan},
  volume    = {162},
  series    = {Proceedings of Machine Learning Research},
  month     = {17--23 Jul},
  publisher = {PMLR},
  url       = {https://proceedings.mlr.press/v162/mindermann22a.html}
}

@inproceedings{ingle-etal-2022-investigating,
  title     = {Investigating the Characteristics of a Transformer in a Few-Shot Setup: Does Freezing Layers in {R}o{BERT}a Help?},
  author    = {Ingle, Digvijay and Tripathi, Rishabh and Kumar, Ayush and Patel, Kevin and Vepa, Jithendra},
  editor    = {Bastings, Jasmijn and Belinkov, Yonatan and Elazar, Yanai and Hupkes, Dieuwke and Saphra, Naomi and Wiegreffe, Sarah},
  booktitle = {Proceedings of the Fifth BlackboxNLP Workshop on Analyzing and Interpreting Neural Networks for NLP},
  month     = dec,
  year      = {2022},
  address   = {Abu Dhabi, United Arab Emirates (Hybrid)},
  publisher = {Association for Computational Linguistics},
  url       = {https://aclanthology.org/2022.blackboxnlp-1.19/},
  doi       = {10.18653/v1/2022.blackboxnlp-1.19},
  pages     = {238--248}
}

@article{Gneiting2007,
  author  = {Gneiting, T. and Raftery, A.E.},
  title   = {Strictly proper scoring rules, prediction, and estimation},
  journal = {Journal of the American Statistical Association},
  year    = {2007},
  volume  = {102},
  number  = {477},
  pages   = {359--378},
  doi     = {10.1198/016214506000001437}
}

@inproceedings{DBLP:conf/csfw/YeomGFJ18,
  author       = {Samuel Yeom and Irene Giacomelli and Matt Fredrikson and Somesh Jha},
  title        = {Privacy Risk in Machine Learning: Analyzing the Connection to Overfitting},
  booktitle    = {31st {IEEE} Computer Security Foundations Symposium, {CSF} 2018, Oxford, United Kingdom, July 9--12, 2018},
  pages        = {268--282},
  publisher    = {{IEEE} Computer Society},
  year         = {2018},
  doi          = {10.1109/CSF.2018.00027},
  url          = {https://doi.org/10.1109/CSF.2018.00027},
  bibsource    = {dblp computer science bibliography, https://dblp.org}
}

@inproceedings{peebles2023scalable,
  title={Scalable diffusion models with transformers},
  author={Peebles, William and Xie, Saining},
  booktitle={Proceedings of the IEEE/CVF international conference on computer vision},
  pages={4195--4205},
  year={2023}
}

@article{lu2019vilbert,
  title={Vilbert: Pretraining task-agnostic visiolinguistic representations for vision-and-language tasks},
  author={Lu, Jiasen and Batra, Dhruv and Parikh, Devi and Lee, Stefan},
  journal={Advances in neural information processing systems},
  volume={32},
  year={2019}
}

@article{kushwaha2023multimodal,
  title={A multimodal prototypical approach for unsupervised sound classification},
  author={Kushwaha, Saksham Singh and Fuentes, Magdalena},
  journal={arXiv preprint arXiv:2306.12300},
  year={2023}
}

@InProceedings{pmlr-v202-zheng23f,
  title = 	 {Revisiting Discriminative vs. Generative Classifiers: Theory and Implications},
  author =       {Zheng, Chenyu and Wu, Guoqiang and Bao, Fan and Cao, Yue and Li, Chongxuan and Zhu, Jun},
  booktitle = 	 {Proceedings of the 40th International Conference on Machine Learning},
  pages = 	 {42420--42477},
  year = 	 {2023},
  editor = 	 {Krause, Andreas and Brunskill, Emma and Cho, Kyunghyun and Engelhardt, Barbara and Sabato, Sivan and Scarlett, Jonathan},
  volume = 	 {202},
  series = 	 {Proceedings of Machine Learning Research},
  month = 	 {23--29 Jul},
  publisher =    {PMLR},
  url = 	 {https://proceedings.mlr.press/v202/zheng23f.html},
}

@article{xue2008,
author = {Xue, Jing-Hao and Titterington, D. Michael},
title = {Comment on "On Discriminative vs. Generative Classifiers: A Comparison of Logistic Regression and Naive Bayes"},
year = {2008},
issue_date = {December  2008},
publisher = {Kluwer Academic Publishers},
address = {USA},
volume = {28},
number = {3},
issn = {1370-4621},
url = {https://doi.org/10.1007/s11063-008-9088-7},
doi = {10.1007/s11063-008-9088-7},
journal = {Neural Process. Lett.},
month = dec,
pages = {169–187},
numpages = {19}
}

@article{sahoo2024simple,
  title={Simple and Effective Masked Diffusion Language Models},
  author={Sahoo, Subham Sekhar and Arriola, Marianne and Schiff, Yair and Gokaslan, Aaron and Marroquin, Edgar and Chiu, Justin T and Rush, Alexander and Kuleshov, Volodymyr},
  journal={arXiv preprint arXiv:2406.07524},
  year={2024}
}

@article{austin2021structured,
  title={Structured denoising diffusion models in discrete state-spaces},
  author={Austin, Jacob and Johnson, Daniel D and Ho, Jonathan and Tarlow, Daniel and Van Den Berg, Rianne},
  journal={Advances in Neural Information Processing Systems},
  volume={34},
  pages={17981--17993},
  year={2021}
}

@article{kumar2020syntax,
    title = "Syntax-Guided Controlled Generation of Paraphrases",
    author = "Kumar, Ashutosh  and
      Ahuja, Kabir  and
      Vadapalli, Raghuram  and
      Talukdar, Partha",
    editor = "Johnson, Mark  and
      Roark, Brian  and
      Nenkova, Ani",
    journal = "Transactions of the Association for Computational Linguistics",
    volume = "8",
    year = "2020",
    address = "Cambridge, MA",
    publisher = "MIT Press",
    url = "https://aclanthology.org/2020.tacl-1.22/",
    doi = "10.1162/tacl_a_00318",
    pages = "329--345",
}

@inproceedings{devlin2019bert,
    title = "{BERT}: Pre-training of Deep Bidirectional Transformers for Language Understanding",
    author = "Devlin, Jacob  and
      Chang, Ming-Wei  and
      Lee, Kenton  and
      Toutanova, Kristina",
    editor = "Burstein, Jill  and
      Doran, Christy  and
      Solorio, Thamar",
    booktitle = "Proceedings of the 2019 Conference of the North {A}merican Chapter of the Association for Computational Linguistics: Human Language Technologies, Volume 1 (Long and Short Papers)",
    month = jun,
    year = "2019",
    address = "Minneapolis, Minnesota",
    publisher = "Association for Computational Linguistics",
    url = "https://aclanthology.org/N19-1423/",
    doi = "10.18653/v1/N19-1423",
    pages = "4171--4186"
}

@inproceedings{kasa-etal-2024-exploring,
    title = "Exploring Ordinality in Text Classification: A Comparative Study of Explicit and Implicit Techniques",
    author = "Kasa, Siva Rajesh  and
      Goel, Aniket  and
      Gupta, Karan  and
      Roychowdhury, Sumegh  and
      Priyatam, Pattisapu  and
      Bhanushali, Anish  and
      Srinivasa Murthy, Prasanna",
    editor = "Ku, Lun-Wei  and
      Martins, Andre  and
      Srikumar, Vivek",
    booktitle = "Findings of the Association for Computational Linguistics: ACL 2024",
    month = aug,
    year = "2024",
    address = "Bangkok, Thailand",
    publisher = "Association for Computational Linguistics",
    url = "https://aclanthology.org/2024.findings-acl.320/",
    doi = "10.18653/v1/2024.findings-acl.320",
    pages = "5390--5404"
}

@article{radford2018improving,
  author = {Radford, Alec and Narasimhan, Karthik and Salimans, Tim and Sutskever, Ilya},
  title = {Improving Language Understanding by Generative Pre-Training},
  year = 2018,
journal = {OpenAI}
}

@inproceedings{
jaini2024intriguing,
title={Intriguing Properties of Generative Classifiers},
author={Priyank Jaini and Kevin Clark and Robert Geirhos},
booktitle={The Twelfth International Conference on Learning Representations},
year={2024},
url={https://openreview.net/forum?id=rmg0qMKYRQ}
}

@inproceedings{
hu2022lora,
title={Lo{RA}: Low-Rank Adaptation of Large Language Models},
author={Edward J Hu and yelong shen and Phillip Wallis and Zeyuan Allen-Zhu and Yuanzhi Li and Shean Wang and Lu Wang and Weizhu Chen},
booktitle={International Conference on Learning Representations},
year={2022},
url={https://openreview.net/forum?id=nZeVKeeFYf9}
}

@inproceedings{he2022diffusionbert,
    title = "{D}iffusion{BERT}: Improving Generative Masked Language Models with Diffusion Models",
    author = "He, Zhengfu  and
      Sun, Tianxiang  and
      Tang, Qiong  and
      Wang, Kuanning  and
      Huang, Xuanjing  and
      Qiu, Xipeng",
    editor = "Rogers, Anna  and
      Boyd-Graber, Jordan  and
      Okazaki, Naoaki",
    booktitle = "Proceedings of the 61st Annual Meeting of the Association for Computational Linguistics (Volume 1: Long Papers)",
    month = jul,
    year = "2023",
    address = "Toronto, Canada",
    publisher = "Association for Computational Linguistics",
    url = "https://aclanthology.org/2023.acl-long.248/",
    doi = "10.18653/v1/2023.acl-long.248",
    pages = "4521--4534"
}

@article{li2022diffusion,
  title={Diffusion-lm improves controllable text generation},
  author={Li, Xiang and Thickstun, John and Gulrajani, Ishaan and Liang, Percy S and Hashimoto, Tatsunori B},
  journal={Advances in neural information processing systems},
  volume={35},
  pages={4328--4343},
  year={2022}
}

@inproceedings{
shi2024simplified,
title={Simplified and Generalized Masked Diffusion for Discrete Data},
author={Jiaxin Shi and Kehang Han and Zhe Wang and Arnaud Doucet and Michalis Titsias},
booktitle={The Thirty-eighth Annual Conference on Neural Information Processing Systems},
year={2024},
url={https://openreview.net/forum?id=xcqSOfHt4g}
}

@inproceedings{socher-etal-2013-recursive,
    title = "Recursive Deep Models for Semantic Compositionality Over a Sentiment Treebank",
    author = "Socher, Richard  and
      Perelygin, Alex  and
      Wu, Jean  and
      Chuang, Jason  and
      Manning, Christopher D.  and
      Ng, Andrew  and
      Potts, Christopher",
    booktitle = "Proceedings of the 2013 Conference on Empirical Methods in Natural Language Processing",
    month = oct,
    year = "2013",
    address = "Seattle, Washington, USA",
    publisher = "Association for Computational Linguistics",
    url = "https://aclanthology.org/D13-1170",
    pages = "1631--1642",
}

@inproceedings{wang-cho-2019-bert,
    title = "{BERT} has a Mouth, and It Must Speak: {BERT} as a {M}arkov Random Field Language Model",
    author = "Wang, Alex  and
      Cho, Kyunghyun",
    editor = "Bosselut, Antoine  and
      Celikyilmaz, Asli  and
      Ghazvininejad, Marjan  and
      Iyer, Srinivasan  and
      Khandelwal, Urvashi  and
      Rashkin, Hannah  and
      Wolf, Thomas",
    booktitle = "Proceedings of the Workshop on Methods for Optimizing and Evaluating Neural Language Generation",
    month = jun,
    year = "2019",
    address = "Minneapolis, Minnesota",
    publisher = "Association for Computational Linguistics",
    url = "https://aclanthology.org/W19-2304/",
    doi = "10.18653/v1/W19-2304",
    pages = "30--36"
}

@article{Efron1975TheEO,
  title={The Efficiency of Logistic Regression Compared to Normal Discriminant Analysis},
  author={Bradley Efron},
  journal={Journal of the American Statistical Association},
  year={1975},
  volume={70},
  pages={892-898},
  url={https://api.semanticscholar.org/CorpusID:34806014}
}

@article{merkle2013choosing,
  title={Choosing a strictly proper scoring rule},
  author={Merkle, Edgar C and Steyvers, Mark},
  journal={Decision Analysis},
  volume={10},
  number={4},
  pages={292--304},
  year={2013},
  publisher={INFORMS}
}

@article{article,
author = {de la Torre, Jordi and Puig, Domenec and Valls, Aida},
year = {2017},
month = {05},
pages = {},
title = {Weighted kappa loss function for multi-class classification of ordinal data in deep learning},
journal = {Pattern Recognition Letters},
doi = {10.1016/j.patrec.2017.05.018}
}

@inproceedings{vaswani2017attention,
 author = {Vaswani, Ashish and Shazeer, Noam and Parmar, Niki and Uszkoreit, Jakob and Jones, Llion and Gomez, Aidan N and Kaiser, \L ukasz and Polosukhin, Illia},
 booktitle = {Advances in Neural Information Processing Systems},
 editor = {I. Guyon and U. Von Luxburg and S. Bengio and H. Wallach and R. Fergus and S. Vishwanathan and R. Garnett},
 pages = {},
 publisher = {Curran Associates, Inc.},
 title = {Attention is All you Need},
 url = {https://proceedings.neurips.cc/paper_files/paper/2017/file/3f5ee243547dee91fbd053c1c4a845aa-Paper.pdf},
 volume = {30},
 year = {2017}
}

@inproceedings{10.5555/2969239.2969312,
author = {Zhang, Xiang and Zhao, Junbo and LeCun, Yann},
title = {Character-level convolutional networks for text classification},
year = {2015},
publisher = {MIT Press},
address = {Cambridge, MA, USA},
booktitle = {Proceedings of the 29th International Conference on Neural Information Processing Systems - Volume 1},
pages = {649–657},
numpages = {9},
location = {Montreal, Canada},
series = {NIPS'15}
}

@inproceedings{saravia-etal-2018-carer,
    title = "{CARER}: Contextualized Affect Representations for Emotion Recognition",
    author = "Saravia, Elvis  and
      Liu, Hsien-Chi Toby  and
      Huang, Yen-Hao  and
      Wu, Junlin  and
      Chen, Yi-Shin",
    booktitle = "Proceedings of the 2018 Conference on Empirical Methods in Natural Language Processing",
    month = oct # "-" # nov,
    year = "2018",
    address = "Brussels, Belgium",
    publisher = "Association for Computational Linguistics",
    url = "https://www.aclweb.org/anthology/D18-1404",
    doi = "10.18653/v1/D18-1404",
    pages = "3687--3697",
}

@inproceedings{maas-etal-2011-learning,
    title = "Learning Word Vectors for Sentiment Analysis",
    author = "Maas, Andrew L.  and
      Daly, Raymond E.  and
      Pham, Peter T.  and
      Huang, Dan  and
      Ng, Andrew Y.  and
      Potts, Christopher",
    editor = "Lin, Dekang  and
      Matsumoto, Yuji  and
      Mihalcea, Rada",
    booktitle = "Proceedings of the 49th Annual Meeting of the Association for Computational Linguistics: Human Language Technologies",
    month = jun,
    year = "2011",
    address = "Portland, Oregon, USA",
    publisher = "Association for Computational Linguistics",
    url = "https://aclanthology.org/P11-1015/",
    pages = "142--150"
}

@inproceedings{pang-lee-2005-seeing,
    title = "Seeing Stars: Exploiting Class Relationships for Sentiment Categorization with Respect to Rating Scales",
    author = "Pang, Bo  and
      Lee, Lillian",
    editor = "Knight, Kevin  and
      Ng, Hwee Tou  and
      Oflazer, Kemal",
    booktitle = "Proceedings of the 43rd Annual Meeting of the Association for Computational Linguistics ({ACL}`05)",
    month = jun,
    year = "2005",
    address = "Ann Arbor, Michigan",
    publisher = "Association for Computational Linguistics",
    url = "https://aclanthology.org/P05-1015/",
    doi = "10.3115/1219840.1219855",
    pages = "115--124"
}

@inproceedings{hateoffensive, title = {Automated Hate Speech Detection and the Problem of Offensive Language}, author = {Davidson, Thomas and Warmsley, Dana and Macy, Michael and Weber, Ingmar}, booktitle = {Proceedings of the 11th International AAAI Conference on Web and Social Media}, series = {ICWSM '17}, year = {2017}, location = {Montreal, Canada}, pages = {512-515} }

@inproceedings{lou2023discrete,
author = {Lou, Aaron and Meng, Chenlin and Ermon, Stefano},
title = {Discrete diffusion modeling by estimating the ratios of the data distribution},
year = {2024},
publisher = {JMLR.org},
booktitle = {Proceedings of the 41st International Conference on Machine Learning},
articleno = {1333},
numpages = {30},
location = {Vienna, Austria},
series = {ICML'24}
}

@article{pattisapu2025leveraging,
  title={Leveraging structural information in tree ensembles for table representation learning},
  author={Pattisapu, Nikhil and Kasa, Siva Rajesh and Roychowdhury, Sumegh and Gupta, Karan and Bhanushali, Anish and Murthy, Prasanna Srinivasa},
  year={2025},
journal={WWW}
}

@article{gupta2023robust,
  title={How robust are LLMs to in-context majority label bias?},
  author={Gupta, Karan and Roychowdhury, Sumegh and Kasa, Siva Rajesh and Kasa, Santhosh Kumar and Bhanushali, Anish and Pattisapu, Nikhil and Murthy, Prasanna Srinivasa},
  journal={arXiv preprint arXiv:2312.16549},
  year={2023}
}

@inproceedings{sun2023text,
  title={Text Classification via Large Language Models},
  author={Sun, Xiaofei and Li, Xiaoya and Li, Jiwei and Wu, Fei and Guo, Shangwei and Zhang, Tianwei and Wang, Guoyin},
  booktitle={Findings of the Association for Computational Linguistics: EMNLP 2023},
  pages={8990--9005},
  year={2023}
}

@article{bickel2009discriminative,
  title={Discriminative learning under covariate shift.},
  author={Bickel, Steffen and Br{\"u}ckner, Michael and Scheffer, Tobias},
  journal={Journal of Machine Learning Research},
  volume={10},
  number={9},
  year={2009}
}

@inproceedings{roychowdhury2024tackling,
  title={Tackling Concept Shift in Text Classification using Entailment-style Modeling},
  author={Roychowdhury, Sumegh and Gupta, Karan and Kasa, Siva Rajesh and Srinivasa Murthy, Prasanna},
  booktitle={Proceedings of the 30th ACM SIGKDD Conference on Knowledge Discovery and Data Mining},
  pages={5647--5656},
  year={2024}
}

@inproceedings{mccallum2006multi,
  title={Multi-conditional learning: Generative/discriminative training for clustering and classification},
  author={McCallum, Andrew and Pal, Chris and Druck, Gregory and Wang, Xuerui},
  booktitle={AAAI},
  volume={1},
  pages={6},
  year={2006}
}

@article{teh2006bayesian,
  title={A Bayesian Interpretation of Interpolated Kneser-Ney NUS School of Computing Technical Report TRA2/06},
  author={Teh, Yee Whye},
  journal={National University of Singapore},
  pages={1--21},
  year={2006}
}

@article{ney1994structuring,
  title={On structuring probabilistic dependences in stochastic language modelling},
  author={Ney, Hermann and Essen, Ute and Kneser, Reinhard},
  journal={Computer Speech \& Language},
  volume={8},
  number={1},
  pages={1--38},
  year={1994},
  publisher={Elsevier}
}

@inproceedings{wuunderstanding2020,
  title={Understanding and Improving Information Transfer in Multi-Task Learning},
  author={Wu, Sen and Zhang, Hongyang R and R{\'e}, Christopher},
  booktitle={International Conference on Learning Representations},
year={2020}
}

@article{hu2023revisiting,
  title={Revisiting scalarization in multi-task learning: A theoretical perspective},
  author={Hu, Yuzheng and Xian, Ruicheng and Wu, Qilong and Fan, Qiuling and Yin, Lang and Zhao, Han},
  journal={Advances in Neural Information Processing Systems},
  volume={36},
  pages={48510--48533},
  year={2023}
}

@ARTICLE{hayashi2021,
  author={Hayashi, Hideaki},
  journal={IEEE Transactions on Neural Networks and Learning Systems}, 
  title={A Hybrid of Generative and Discriminative Models Based on the Gaussian-Coupled Softmax Layer}, 
  year={2025},
  volume={36},
  number={2},
  pages={2894-2904},
  doi={10.1109/TNNLS.2024.3358113}}

@article{raina2003classification,
  title={Classification with hybrid generative/discriminative models},
  author={Raina, Rajat and Shen, Yirong and Mccallum, Andrew and Ng, Andrew},
  journal={Advances in neural information processing systems},
  volume={16},
  year={2003}
}

@article{blasiok2023does,
  title={When does optimizing a proper loss yield calibration?},
  author={Blasiok, Jaroslaw and Gopalan, Parikshit and Hu, Lunjia and Nakkiran, Preetum},
  journal={Advances in Neural Information Processing Systems},
  volume={36},
  pages={72071--72095},
  year={2023}
}

@inproceedings{quinonero2005evaluating,
  title={Evaluating predictive uncertainty challenge},
  author={Quinonero-Candela, Joaquin and Rasmussen, Carl Edward and Sinz, Fabian and Bousquet, Olivier and Sch{\"o}lkopf, Bernhard},
  booktitle={Machine Learning Challenges Workshop},
  pages={1--27},
  year={2005},
  organization={Springer}
}

@article{vincent2010stacked,
  title={Stacked denoising autoencoders: Learning useful representations in a deep network with a local denoising criterion.},
  author={Vincent, Pascal and Larochelle, Hugo and Lajoie, Isabelle and Bengio, Yoshua and Manzagol, Pierre-Antoine and Bottou, L{\'e}on},
  journal={Journal of machine learning research},
  volume={11},
  number={12},
  year={2010}
}

@article{rezaee2021discriminative,
  title={Discriminative and generative transformer-based models for situation entity classification},
  author={Rezaee, Mehdi and Darvish, Kasra and Kebe, Gaoussou Youssouf and Ferraro, Francis},
  journal={arXiv preprint arXiv:2109.07434},
  year={2021}
}

@inproceedings{li2019generative,
  title={Are generative classifiers more robust to adversarial attacks?},
  author={Li, Yingzhen and Bradshaw, John and Sharma, Yash},
  booktitle={International Conference on Machine Learning},
  pages={3804--3814},
  year={2019},
  organization={PMLR}
}

@inproceedings{li2024generative,
  title={Generative Classifiers Avoid Shortcut Solutions},
  author={Li, Alexander Cong and Kumar, Ananya and Pathak, Deepak},
  booktitle={The Thirteenth International Conference on Learning Representations},
  year={2025}
}

@inproceedings{stanley2025does,
  title={Does a diffusion-based generative classifier avoid shortcut learning in medical image analysis? An initial investigation using synthetic neuroimaging data},
  author={Stanley, Emma AM and Forkert, Nils D and Wilms, Matthias},
  booktitle={Medical Imaging 2025: Imaging Informatics},
  volume={13411},
  pages={94--99},
  year={2025},
  organization={SPIE}
}

@article{zeng2023certified,
  title={Certified robustness to text adversarial attacks by randomized [mask]},
  author={Zeng, Jiehang and Xu, Jianhan and Zheng, Xiaoqing and Huang, Xuanjing},
  journal={Computational Linguistics},
  volume={49},
  number={2},
  pages={395--427},
  year={2023},
  publisher={MIT Press One Broadway, 12th Floor, Cambridge, Massachusetts 02142, USA~…}
}

@article{razeghi2022impact,
  title={Impact of pretraining term frequencies on few-shot reasoning},
  author={Razeghi, Yasaman and Logan IV, Robert L and Gardner, Matt and Singh, Sameer},
  journal={arXiv preprint arXiv:2202.07206},
  year={2022}
}

@inproceedings{zhang2024text,
  title={Text-crs: A generalized certified robustness framework against textual adversarial attacks},
  author={Zhang, Xinyu and Hong, Hanbin and Hong, Yuan and Huang, Peng and Wang, Binghui and Ba, Zhongjie and Ren, Kui},
  booktitle={2024 IEEE Symposium on Security and Privacy (SP)},
  pages={2920--2938},
  year={2024},
  organization={IEEE}
}

@inproceedings{erhan2010does,
  title={Why does unsupervised pre-training help deep learning?},
  author={Erhan, Dumitru and Courville, Aaron and Bengio, Yoshua and Vincent, Pascal},
  booktitle={Proceedings of the thirteenth international conference on artificial intelligence and statistics},
  pages={201--208},
  year={2010},
  organization={JMLR Workshop and Conference Proceedings}
}

@article{wang2023calibration,
  title={Calibration in deep learning: A survey of the state-of-the-art},
  author={Wang, Cheng},
  journal={arXiv preprint arXiv:2308.01222},
  year={2023}
}

@InProceedings{pmlr-v70-guo17a,
  title = 	 {On Calibration of Modern Neural Networks},
  author =       {Chuan Guo and Geoff Pleiss and Yu Sun and Kilian Q. Weinberger},
  booktitle = 	 {Proceedings of the 34th International Conference on Machine Learning},
  pages = 	 {1321--1330},
  year = 	 {2017},
  editor = 	 {Precup, Doina and Teh, Yee Whye},
  volume = 	 {70},
  series = 	 {Proceedings of Machine Learning Research},
  month = 	 {06--11 Aug},
  publisher =    {PMLR},
  url = 	 {https://proceedings.mlr.press/v70/guo17a.html}
}

@inproceedings{he2016deep,
  title={Deep residual learning for image recognition},
  author={He, Kaiming and Zhang, Xiangyu and Ren, Shaoqing and Sun, Jian},
  booktitle={Proceedings of the IEEE conference on computer vision and pattern recognition},
  pages={770--778},
  year={2016}
}

@article{huang2019clinicalbert,
  title={Clinicalbert: Modeling clinical notes and predicting hospital readmission},
  author={Huang, Kexin and Altosaar, Jaan and Ranganath, Rajesh},
  journal={arXiv preprint arXiv:1904.05342},
  year={2019}
}

@inproceedings{martin-etal-2022-swahbert,
    title = "{S}wah{BERT}: Language Model of {S}wahili",
    author = "Martin, Gati  and
      Mswahili, Medard Edmund  and
      Jeong, Young-Seob  and
      Woo, Jiyoung",
    editor = "Carpuat, Marine  and
      de Marneffe, Marie-Catherine  and
      Meza Ruiz, Ivan Vladimir",
    booktitle = "Proceedings of the 2022 Conference of the North American Chapter of the Association for Computational Linguistics: Human Language Technologies",
    month = jul,
    year = "2022",
    address = "Seattle, United States",
    publisher = "Association for Computational Linguistics",
    url = "https://aclanthology.org/2022.naacl-main.23/",
    doi = "10.18653/v1/2022.naacl-main.23",
    pages = "303--313"
}

@article{liu2019roberta,
  title={Roberta: A robustly optimized bert pretraining approach},
  author={Liu, Yinhan and Ott, Myle and Goyal, Naman and Du, Jingfei and Joshi, Mandar and Chen, Danqi and Levy, Omer and Lewis, Mike and Zettlemoyer, Luke and Stoyanov, Veselin},
  journal={arXiv preprint arXiv:1907.11692},
  year={2019}
}

@inproceedings{li2022diffusionlm,
  title={Diffusion-LM Improves Controllable Text Generation},
  author={Li, Xueguang and others},
  booktitle={Advances in Neural Information Processing Systems (NeurIPS)},
  year={2022}
}

@article{sun2022score,
  title={Score-based continuous-time discrete diffusion models},
  author={Sun, Haoran and Yu, Lijun and Dai, Bo and Schuurmans, Dale and Dai, Hanjun},
  journal={arXiv preprint arXiv:2211.16750},
  year={2022}
}

@ARTICLE{6797087,
  author={Geman, Stuart and Bienenstock, Elie and Doursat, René},
  journal={Neural Computation}, 
  title={Neural Networks and the Bias/Variance Dilemma}, 
  year={1992},
  volume={4},
  number={1},
  pages={1-58},
  doi={10.1162/neco.1992.4.1.1}}

@misc{nakkiran2019deepdoubledescentbigger,
      title={Deep Double Descent: Where Bigger Models and More Data Hurt}, 
      author={Preetum Nakkiran and Gal Kaplun and Yamini Bansal and Tristan Yang and Boaz Barak and Ilya Sutskever},
      year={2019},
      eprint={1912.02292},
      archivePrefix={arXiv},
      primaryClass={cs.LG},
      url={https://arxiv.org/abs/1912.02292}, 
}

@article{Belkin_2019,
   title={Reconciling modern machine-learning practice and the classical bias–variance trade-off},
   volume={116},
   ISSN={1091-6490},
   url={http://dx.doi.org/10.1073/pnas.1903070116},
   DOI={10.1073/pnas.1903070116},
   number={32},
   journal={Proceedings of the National Academy of Sciences},
   publisher={Proceedings of the National Academy of Sciences},
   author={Belkin, Mikhail and Hsu, Daniel and Ma, Siyuan and Mandal, Soumik},
   year={2019},
   month=jul, pages={15849–15854} }

\newpage
\appendix

\section{More Background and Related works}
\label{app:background_related_works}
\textbf{Discrete Diffusion Models for Classification.}
Recent advances in discrete diffusion models have shown promising results in text generation tasks, matching or surpassing autoregressive models at GPT-2 scale \citep{lou2023discrete,sahoo2024simple,shi2024simplified}. While these models have demonstrated success in controlled generation tasks \citep{li2022diffusion, he2022diffusionbert}, specifically syntax controlled generation of text \cite{kumar2020syntax} and text infilling, their application to classification remains relatively unexplored. Traditional diffusion models for text generation, such as DiffusionBERT~\citep{he2022diffusionbert}, DiffusionLM~\citep{li2022diffusionlm}, and D3PM~\citep{austin2021structured}, operate by embedding discrete token sequences into continuous spaces and applying Gaussian noise-based diffusion. In contrast, SEDD~\citep{lou2023discrete} was the first to directly model diffusion in discrete space through a score entropy-driven objective. Hence, we adopt SEDD as our baseline method.
Our work provides the first systematic evaluation of discrete diffusion models for classification tasks, comparing them against traditional discriminative and generative approaches.

\noindent \textbf{Robustness to Noise.}
Previous studies have examined robustness primarily through the lens of adversarial attacks \cite{li2019generative}, distribution shifts \cite{li2024generative} and domain shifts \cite{jaini2024intriguing}. While recent work has provided certified robustness guarantees for perturbations like insertion, deletion, reordering and synonyms for specific architectures \cite{zeng2023certified,zhang2024text}, our study presents comparisons across model families under two different noise conditions in the context of TC for transformer architectures.

\noindent \textbf{Calibration \& Ordinality}. 
Model calibration is crucial in classification, as it reflects how well predicted probabilities align with actual frequencies. Proper Scoring Rules (PSR) \cite{merkle2013choosing} offer a theoretical basis for producing calibrated predictions: a scoring rule (i.e. loss function) is proper if its expected value is minimized only when predicted probabilities match the true distribution. All our modeling approaches—Generative (AR, MLM, Discrete Diffusion) and Discriminative (Encoder)—optimize proper scoring rules, but only \texttt{ENC} demonstrates consistently calibrated results because it optimizes a loss directly aligned with the classification task. Although the other paradigms also optimize strictly proper scoring rules that guarantee the lowest expected score when predictions match the target distribution, they optimize different objectives that do not perfectly align with classification. This mismatch in optimization targets explains the differences observed in calibration performance. GPT and MLM maximize likelihood, Discrete Diffusion optimizes a variational bound, and cross-entropy minimizes the KL-divergence between predicted and true distributions. Recent work~\cite{blasiok2023does} shows that models trained with PSRs are often naturally calibrated when achieving low training loss, without requiring post-hoc calibration. This motivates us to empirically assess calibration across our models, as their differing architectures and objectives may still lead to varying calibration behaviors.

Ordinality in text classification is essential for applications like sentiment analysis or medical assessments, where label order affects decisions and distant misclassifications are more harmful. Recent works \cite{kasa-etal-2024-exploring} systematically compare \emph{explicit} methods—like custom losses enforcing label order—with \emph{implicit} approaches using pretrained models’ semantics. However, no prior work focuses on exploring ordinality across diverse modeling frameworks trained from scratch.

\section{Implementation Details}
\label{app:implementation}

We use the \texttt{bert-base-uncased}\footnote{\url{https://huggingface.co/google-bert/bert-base-uncased}} architecture as the backbone for our \textbf{Encoder} and \textbf{MLM} experiments, without initializing the model with pretrained weights. This architecture contains approximately 110M parameters, comprising 12 encoder layers, 12 attention heads, and a hidden size of 768. We run all experiments for 3 random seeds and report the average and standard deviation results in main paper.

For the \textbf{Encoder} experiments, we conducted a grid search over several hyperparameters, including learning rates of \{\texttt{1e-5}, \texttt{2e-5}, \texttt{3e-5}, \texttt{4e-5}, \texttt{5e-5}\}, batch sizes of \{32, 64, 128, 256\}, and a fixed sequence length of 512 tokens. Training was performed for 30 epochs uniformly for all datasets without early stopping. For the \textbf{MLM}-based experiments, we retained similar hyperparameter ranges but trained for 200 epochs to account for the increased complexity of masked token prediction. We observed that adding an early stopping patience parameter sometimes led the model to select a suboptimal checkpoint, as the validation loss often continued to decrease gradually after remaining flat or oscillating for several epochs.

For the \textbf{AR} and \textbf{AR$_{pseudo}$} experiments, we used the \texttt{GPT-2} base architecture\footnote{\url{https://huggingface.co/openai-community/gpt2}} as the backbone with 137M parameters comparable with our other experiments. We trained a causal language model to minimize the next-token prediction loss over the concatenated input and label sequence. A grid search was conducted with the same hyperparameter range as mentioned above. The models were trained for up to 100 epochs, with early stopping based on validation loss, using a patience parameter of 10 epochs.

Our \textbf{Text Diffusion} approach follows the Diffusion Transformer architecture \cite{peebles2023scalable} which is basically the vanilla transformer encoder with an extra time-conditioned embedding incorporated with it. The parameter count is $\sim$160M due to the addition of time-dependent embeddings required by the diffusion mechanism. To counter this, we conducted an ablation study by increasing the encoder size to 160M parameters (by adding layers) for other approaches (like \texttt{ENC}, \texttt{MLM}) to match the diffusion model size, but observed no difference in performance. Hence we retain their original settings as reported above.
For diffusion-specific hyperparameters, we used a batch size of 64, learning rate \texttt{3e-4} and trained for 200K iterations. We adopted a geometric noise schedule that interpolates between $10^{-4}$ and 20, similar to the setup in \cite{lou2023discrete}, and used the following absorbing/masking matrix $Q^{absorb}$ as part of the transition modeling. This was the best hyperparameter setting we found.
\[
Q_{\text{absorb}} = 
\begin{bmatrix}
-1 & 0 & \cdots & 0 \\
0 & -1 & \cdots & 0 \\
\vdots & \vdots & \ddots & \vdots \\
1 & 1 & \cdots &0
\end{bmatrix}
\]

All experiments were conducted using multi-GPU training across eight NVIDIA A100 GPUs. Training time varied depending on the methods and configurations used for each dataset. The range of training times (in hours) for various datasets is presented in Table \ref{tab:training_time}. All reported training times correspond to full-data training configurations. 

Our analysis of inference latency reveals significant differences across architectures - refer to Table \ref{tab:inference_latency}. While ENC and MLM demonstrate comparable inference speeds (requiring single forward passes), AR requires |K| forward passes for prediction, though this can be parallelized at the cost of increased computation. DIFF exhibits substantially higher latency, taking approximately 20-100x longer than ENC/MLM due to its iterative denoising process. Specifically, for a batch of 1024 examples (sequence length 128) on an A100 GPU, ENC and MLM take ~0.03s for small models (3.3M params) to ~1.3s for large models (120.4M params), while DIFF requires 16-25s across model sizes.

\begin{figure}[t]
    \centering

    \setlength{\tabcolsep}{4pt}
    \resizebox{\columnwidth}{!}{%
    \begin{tabular}{@{}lccccc@{}}
        \toprule
        \textbf{Config} & ENC & AR$_{pseudo}$ & AR & MLM & DIFF \\
        \midrule
        (1L,1H) &  1-2 & 2-4 & 2-4 & 1-4 & 1-4 \\
        (6L,6H)  & 1-3 & 3-7 & 3-7 & 3-7 & 2-6 \\
        (12L,12H)  &  2-5 & 5-10 & 5-10 & 5-10 & 5-12 \\
        \bottomrule
    \end{tabular}%
    }
    \captionof{table}{Training time (in hrs) ranges across different datasets for each configuration and approach.}
    \label{tab:training_time}
\end{figure}

\begin{table}[htp!]
\centering
\resizebox{\columnwidth}{!}{%
\begin{tabular}{lccccc}
\toprule
\textbf{Model Size} & Parameters & ENC & MLM & AR & DIFF \\
\midrule
Small  & 3.3M   & \cellcolor{lightgreen!50}0.027 & \cellcolor{lightgreen!50}0.027 & \cellcolor{lightyellow!50}0.058 & 16.2  \\
Medium & 30.3M  & \cellcolor{lightgreen!50}0.292 & \cellcolor{lightgreen!50}0.292 & \cellcolor{lightyellow!50}0.510 & 20.52 \\
Large  & 120.4M & \cellcolor{lightgreen!50}1.260 & \cellcolor{lightgreen!50}1.260 & \cellcolor{lightyellow!50}2.070 & 24.8  \\
\bottomrule
\end{tabular}%
}
\caption{Model Size v/s Inference Latency (avg wall-clock time per batch in seconds)}
\label{tab:inference_latency}
\end{table}

\section{Dataset Details}
\label{app:datasets}
\begin{table*}[htp!]
\small
\centering
\begin{tabular}{lllclp{2.9cm}c}
\toprule
\textbf{Dataset} & \textbf{Split} & \textbf{Examples} & \textbf{Classes} & \textbf{Avg Tokens} & \textbf{Label Dist. (\%)} & \textbf{Ordinal}\\
\toprule
\multirow{2}{*}{IMDb} & train & 25,000 & 2 & 313.87 & 0: 50.0, 1: 50.0 & $\times$\\
& test & 25,000 & 2 & 306.77  & 0: 50.0, 1: 50.0 \\
\midrule
\multirow{2}{*}{agnews} & train & 120,000 & 4 & 53.17  & 0-3: 25.0 each & $\times$\\
& test & 7,600 & 4 & 52.75 & 0-3: 25.0 each \\
\midrule
\multirow{2}{*}{emotion} & train & 16,000 & 6 & 22.26  & 0: 29.2, 1: 33.5, 2: 8.2, 3: 13.5, 4: 12.1, 5: 3.6 & $\times$\\
& test & 2,000 & 6 & 21.90  & 0: 27.5, 1: 35.2, 2: 8.9, 3: 13.8, 4: 10.6, 5: 4.1 \\
\midrule
\multirow{2}{*}{hatespeech} & train & 22,783 & 3 & 30.04  & 0: 5.8, 1: 77.5, 2: 16.7 & \checkmark \\
& test & 2,000 & 3 & 30.18 & 0: 5.5, 1: 76.6, 2: 17.9 \\
\midrule
\multirow{2}{*}{multiclasssentiment} & train & 31,232 & 3 & 26.59 & 0: 29.2, 1: 37.3, 2: 33.6 & \checkmark \\
& test & 5,205 & 3 & 26.91 & 0: 29.2, 1: 37.0, 2: 33.8 \\
\midrule
\multirow{2}{*}{rottentomatoes} & train & 8,530 & 2 & 27.37 & 0: 50.0, 1: 50.0 & $\times$\\
& test & 1,066 & 2 & 27.32 & 0: 50.0, 1: 50.0 \\
\midrule
\multirow{2}{*}{sst2} & train & 6,920 & 2 & 25.21  & 0: 47.8, 1: 52.2 & $\times$ \\
& test & 872 & 2 & 25.47 & 0: 49.1, 1: 50.9 \\
\midrule
\multirow{2}{*}{sst5} & train & 8,544 & 5 & 25.04  & 0: 12.8, 1: 26.0, 2: 19.0, 3: 27.2, 4: 15.1  & \checkmark\\
& test & 1,101 & 5 & 25.24  & 0: 12.6, 1: 26.3, 2: 20.8, 3: 25.3, 4: 15.0 \\
\midrule
\multirow{2}{*}{twitter} & train & 9,543 & 3 & 27.62 & 0: 15.1, 1: 20.2, 2: 64.7  & \checkmark\\
& test & 2,388 & 3 & 27.92  & 0: 14.5, 1: 19.9, 2: 65.6 \\
\bottomrule
\end{tabular}
\caption{Dataset statistics showing training and test split sizes, number of classes, mean and maximum token lengths, and label distribution percentages. Refer to \refsec{app:datasets} for details on datasets.}
\label{tab:dataset_stats}
\end{table*}
\textbf{AG News} \cite{10.5555/2969239.2969312}: It consists of approximately 120K training samples and 7.6K test samples, divided into four categories: World, Sports, Business, and Technology. Each sample contains a short news article, typically consisting of the title and the first few sentences. \textbf{Emotion} \cite{saravia-etal-2018-carer}: A collection of English tweets labeled with six basic emotions: anger, fear, joy, love, sadness, and surprise. It is designed for emotion detection in text. The dataset has 20K samples divided into 16K samples for training and 2K samples each for validation and testing. \textbf{Stanford Sentiment Treebank (SST)} \cite{socher-etal-2013-recursive}: We utilize both the SST-2 (binary sentiment) and SST-5 (fine-grained sentiment) variants of the Stanford Sentiment Treebank dataset. SST-2 consists of sentences labeled as either positive or negative, suitable for binary sentiment classification, while SST-5 includes five sentiment categories: very negative, negative, neutral, positive, and very positive, allowing for more fine-grained sentiment analysis.
\textbf{Multiclass Sentiment Analysis} \footnote{https://huggingface.co/datasets/Sp1786/multiclass-sentiment-analysis-dataset}:  This dataset consists of 41.6K data points, labeled into three sentiment categories: positive, negative, and neutral. While the dataset is designed for multiclass sentiment classification, it exhibits class imbalance, with certain sentiment classes being more prevalent than others. This imbalance provides a more realistic challenge for sentiment analysis models, testing their ability to handle skewed distributions and still perform effectively across all sentiment categories.
\textbf{Twitter Financial News Sentiment} \footnote{https://huggingface.co/datasets/zeroshot/twitter-financial-news-sentiment}: A specialized English-language collection of finance-related tweets, annotated for sentiment analysis. It consists of 11,932 tweets labeled with three sentiment categories: Bearish, Bullish, and Neutral. This dataset is designed to test models' ability to understand domain-specific language and nuanced sentiment expressions in financial contexts.
\textbf{IMDb} \cite{maas-etal-2011-learning}: A binary sentiment analysis dataset consisting of 50K reviews from the Internet Movie Database (IMDb), labeled as positive or negative. The dataset is balanced, with an equal number of positive and negative reviews. This dataset is characterized by longer document lengths and detailed opinions, making it a challenging benchmark.
\textbf{Rotten Tomatoes} \cite{pang-lee-2005-seeing}: A binary classification dataset which contains 10,662 movie review sentences, equally divided into 5,331 positive and 5,331 negative examples. The dataset is characterized by relatively short, opinion-driven sentences that reflect concise sentiments about films.
\textbf{Hate Speech Offensive} \cite{hateoffensive}: A major challenge in automatic hate speech detection is distinguishing hate speech from other forms of offensive language. This dataset consists of approximately 25K tweets, labeled into three categories: hate speech, offensive language without hate speech, and neutral content.

Refer to Table \ref{tab:dataset_stats} for details on dataset statistics.

\FloatBarrier
\onecolumn
\section{More Ordinal \& Calibration Results}
\label{app:ordinal_calibration}
In this section, we take a closer look at ordinal and calibration results for the datasets decribed above. Here we report ordinal metrics on the datasets \textbf{Stanford Sentiment Treebank (SST5)} \cite{socher-etal-2013-recursive}, \textbf{Multiclass Sentiment Analysis}, \textbf{Hate Speech Offensive} \cite{hateoffensive} and
\textbf{Twitter Financial News Sentiment} since these are the only multi-class ordinal datasets out of 9. Calibration metrics are reported on all 9 datasets. 

\begin{figure}[h]
    \centering
    \includegraphics[width=1\textwidth]{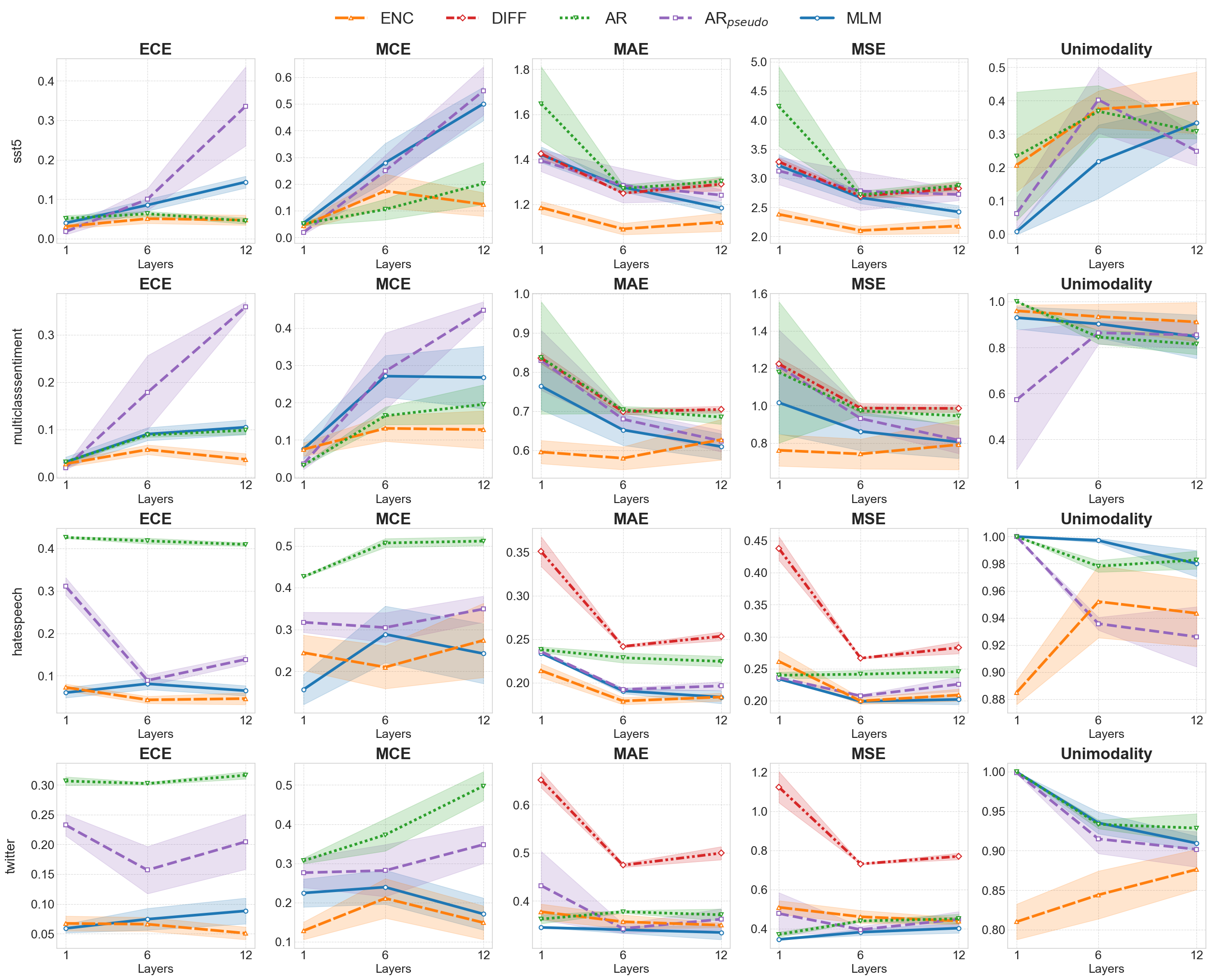}
    \caption{\small\textbf{[Best viewed in color]} Calibration and Ordinal metrics comparison across layers 1, 6 and 12. For ECE, MCE, MAE, MSE, ($\downarrow$ is better) and UM ($\uparrow$ is better).}
    \label{fig:calibration_ordinal_across_layers}
\end{figure}

\begin{figure}[p]
    \centering
    \includegraphics[width=1\textwidth]{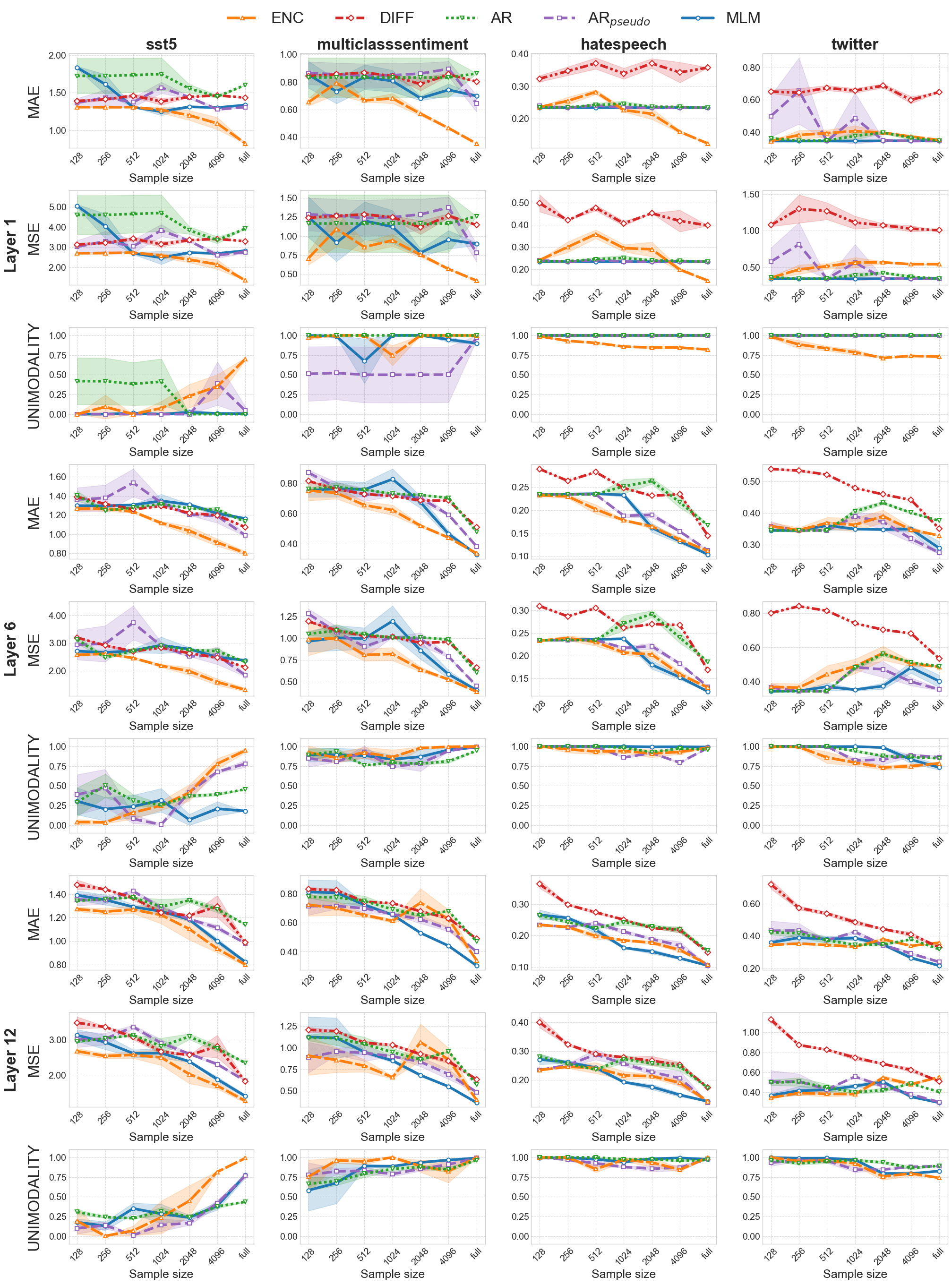}
    \caption{\small\textbf{[Best viewed in color]} Ordinal metrics. For MAE, MSE, ($\downarrow$ is better) and UM ($\uparrow$ is better).}
    \label{fig:ordinal_metrics_layers_6}
\end{figure}

\begin{figure}[p]
    \centering
    \includegraphics[width=1\textwidth]{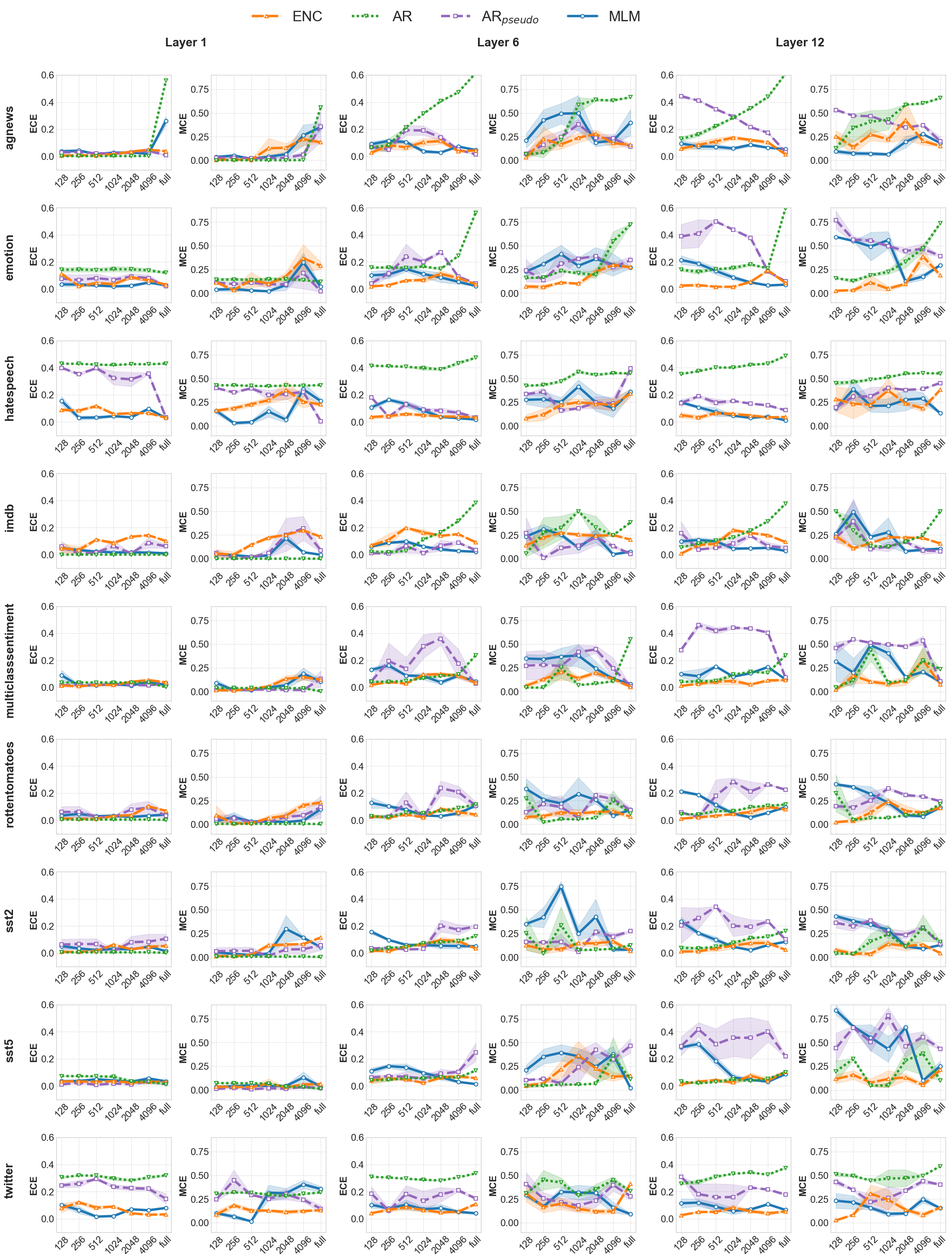}
    \caption{\small\textbf{[Best viewed in color]} Calibration metrics. For ECE, MCE ($\downarrow$ is better)}
    \label{fig:calibration_9_datasets_layers_6}
\end{figure}

In Figure~\ref{fig:calibration_ordinal_across_layers}, we compare how ordinal and calibration metrics vary with increasing model size. Figure~\ref{fig:ordinal_metrics_layers_6} presents the ordinal metrics for all four ordinal datasets, while Figure~\ref{fig:calibration_9_datasets_layers_6} shows the calibration metrics for all nine datasets. The corresponding insights are discussed in Section~\ref{sec:results} (see \textbf{Q3}).

\FloatBarrier

\clearpage

\clearpage

\clearpage

\FloatBarrier 
\clearpage
\newpage

\section{More Main Results}
\label{app:main_results}
\begin{figure}[H]
    \centering

\includegraphics[width=\textwidth]{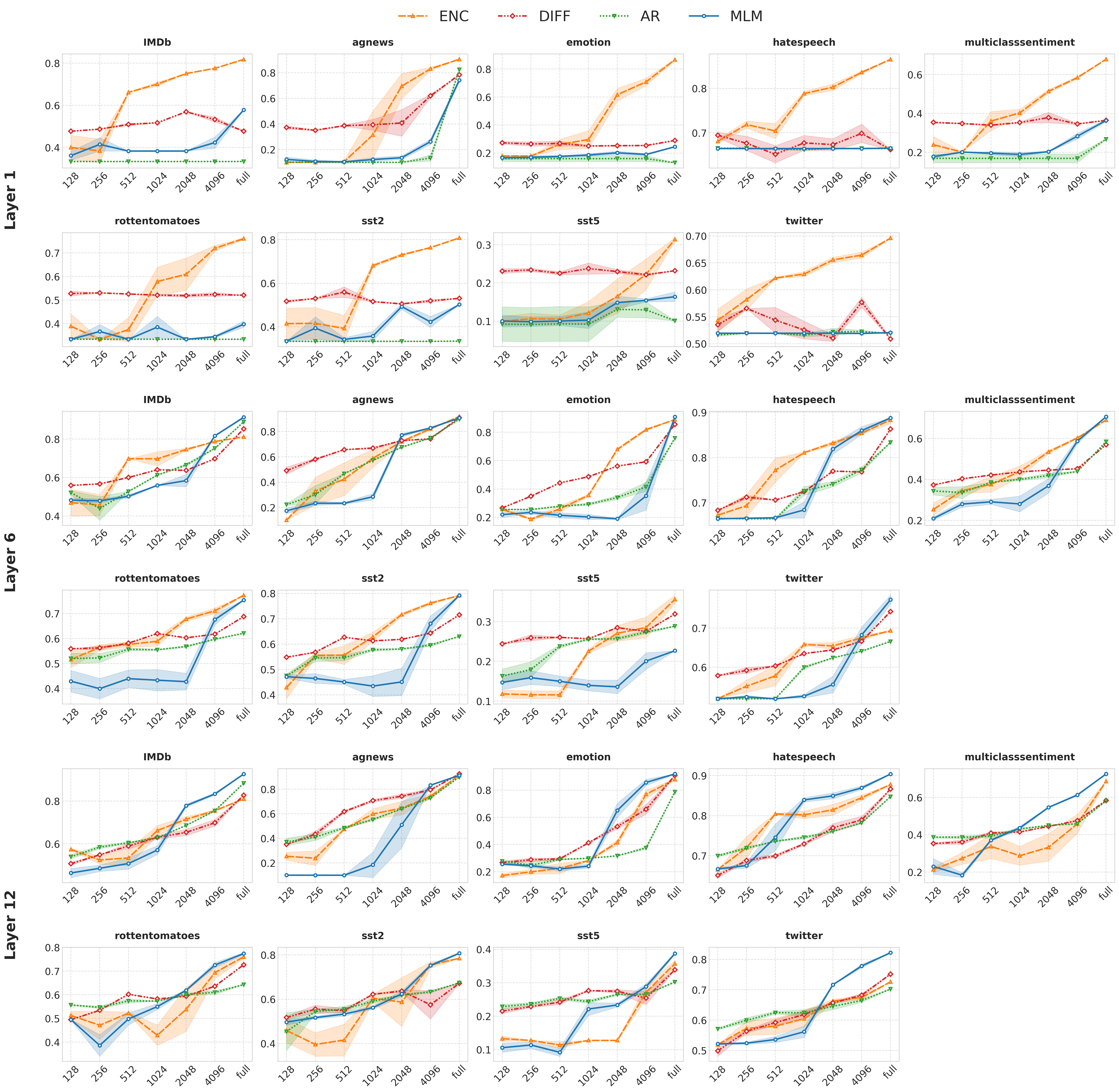}
    
    \caption{\small \textbf{[Best viewed in color]} Comparison of weighted-F1 scores of models across different configurations for all 9 datasets. ($\uparrow$ is better) (\texttt{X-axis}: sample size, \texttt{Y-axis}: weighted-F1 score)}

    \label{fig:combined_plots_all_data}
\end{figure}

\begin{figure*}[h]
    \centering

\includegraphics[width=\textwidth]{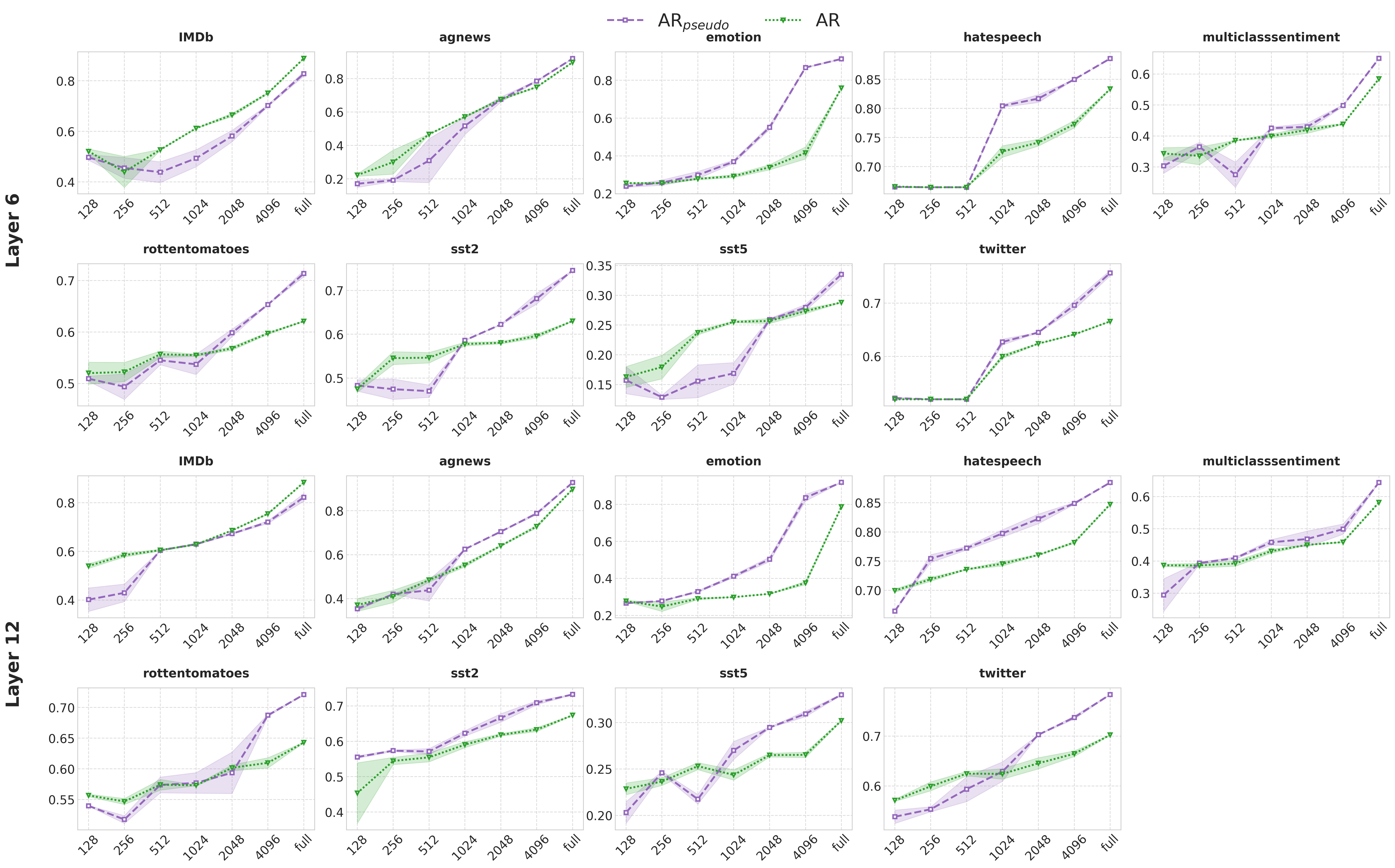}
    
    \caption{\small \textbf{[Best viewed in color]} Comparison of weighted-F1 scores between \texttt{AR$_{pseudo}$} and \texttt{AR} ($\uparrow$ is better) for all datasets. (\texttt{X-axis}: sample size, \texttt{Y-axis}: weighted-F1 score)}

    \label{fig:combined_plots_gpts_all_data}
\end{figure*}

This section contains the extended results of Figure~\ref{fig:combined_plots} (see Figure~\ref{fig:combined_plots_all_data}) and Figure~\ref{fig:combined_plots_gpts} (see Figure~\ref{fig:combined_plots_gpts_all_data}) for all 9 datasets. We omit 1-layer plots for Figure~\ref{fig:combined_plots_gpts_all_data} since the performance is mostly trivial for low-data settings and the same trend is observed as 6/12-layers for full-data settings.

\section{Results on Pretrained models }
\label{sec:pretrained_results}

\begin{figure*}[h]
    \centering

\includegraphics[width=\textwidth]{sections/figures/ar_enc_datasets_cameraready_top_margin.png}
    
    \caption{\small \textbf{[Best viewed in color]} Comparison of weighted-F1 scores between pretrained \texttt{AR} and \texttt{ENC} models ($\uparrow$ is better). (\texttt{X-axis}: sample size, \texttt{Y-axis}: weighted-F1 score)}

    \label{fig:pretrained_plot}
\end{figure*}

\newpage
\section{Revisiting the Bias-Variance Trade-off in Modern Text Classifiers}
\label{sec:revisiting_bias_variance}

\begin{figure*}[h]
    \centering

\includegraphics[width=\textwidth]{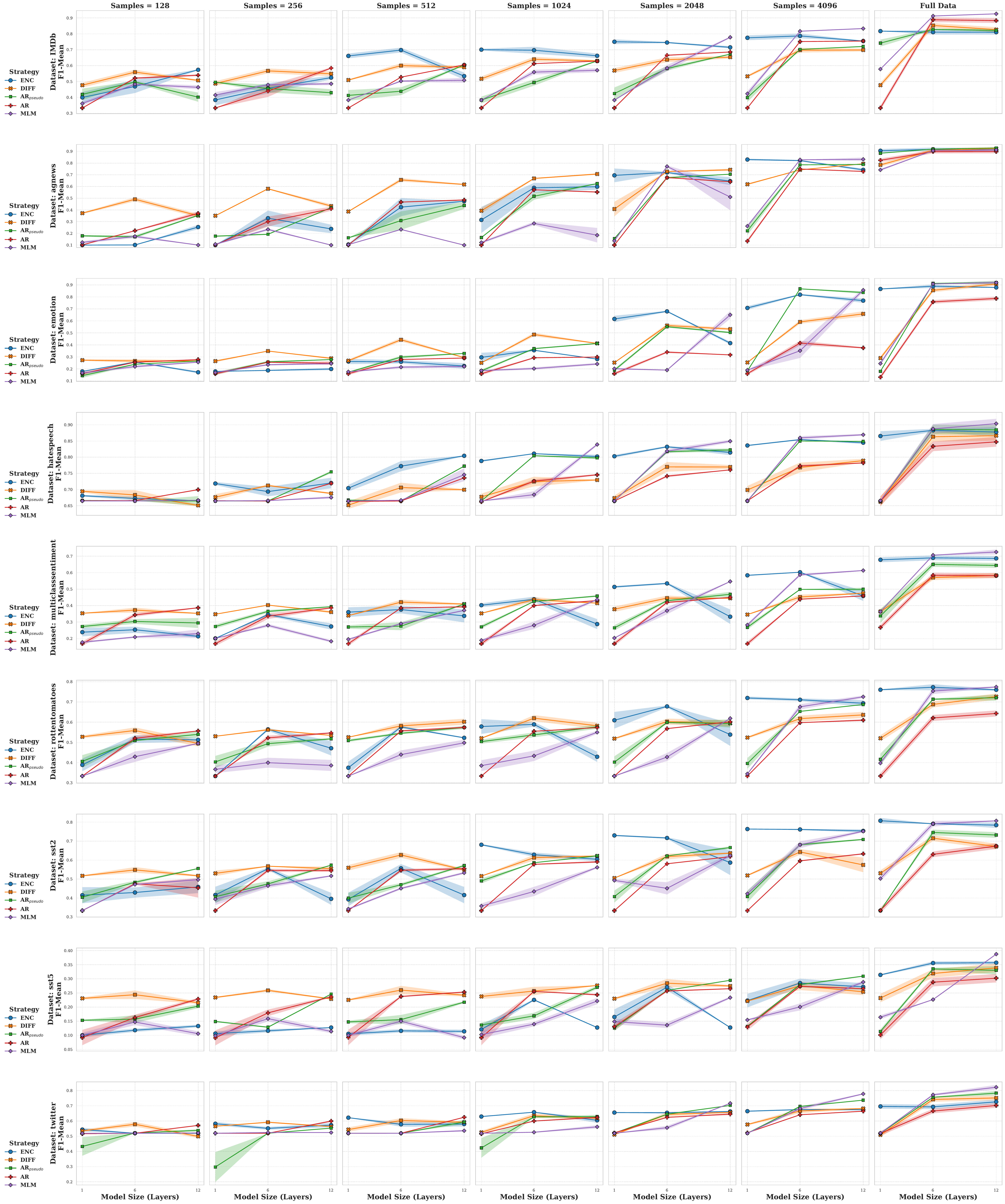}
    
    \caption{\small \textbf{[Best viewed in color]} Comparison of weighted F1-scores between training strategies ENC, DIFF, AR, \(AR_{\text{pseudo}}\), and MLM across different sample sizes as model size increases. The plots highlight that the classical bias-variance trade-off phenomenon is only evident in small-data settings (i.e., lower sample sizes).}

    \label{fig:iterate_over_datasets}
\end{figure*}

\begin{figure*}[h]
    \centering

\includegraphics[width=0.77\textwidth]{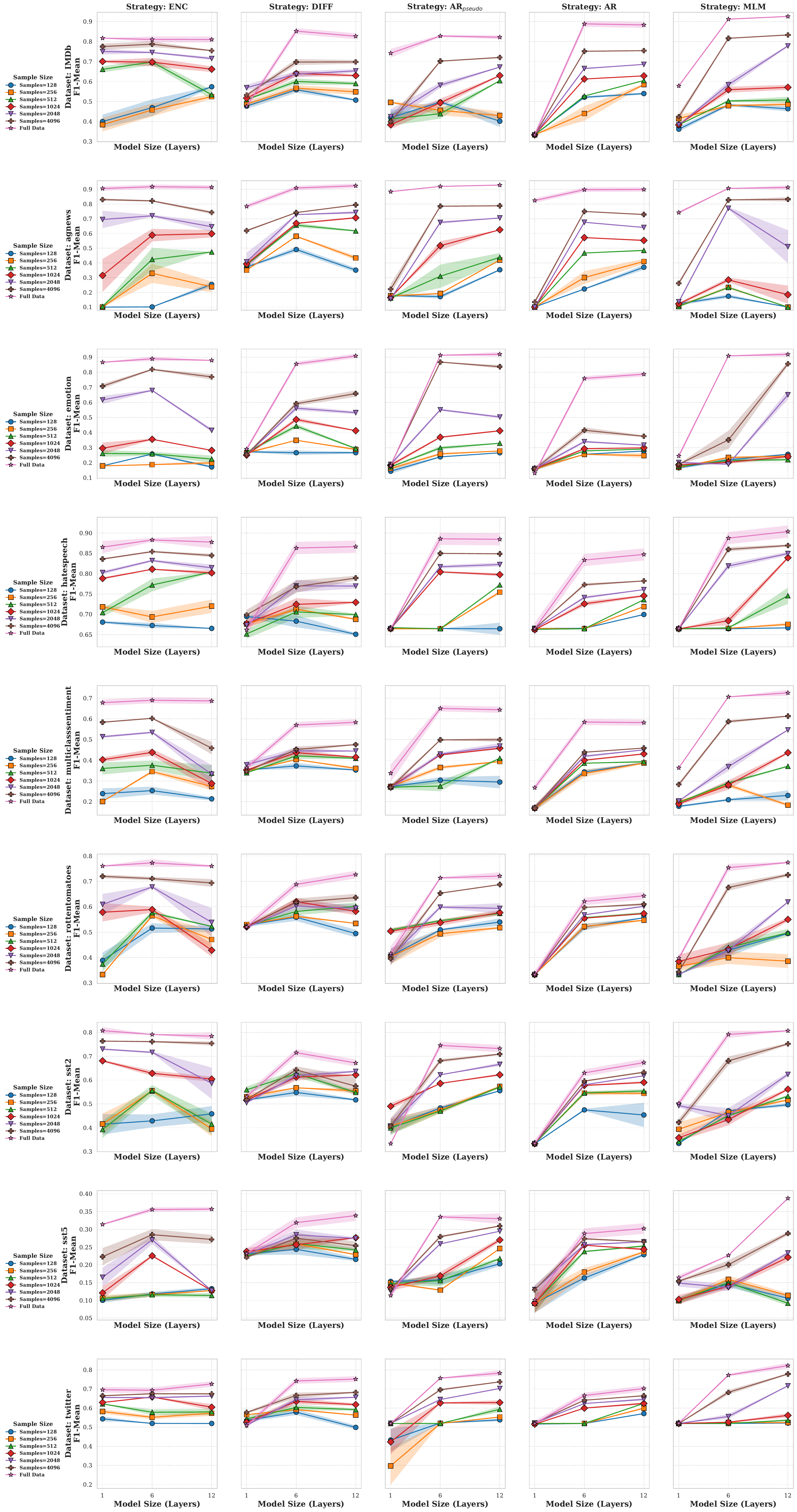}
    
    \caption{\small \textbf{[Best viewed in color]} Comparison of F1-Mean scores across different sample sizes and strategies as model size increases. The figure highlights that, in some cases, increasing training data can adversely affect performance.}

    \label{fig:iterate_over_strategy}
\end{figure*}

The classical bias-variance trade-off, which predicts a U-shaped performance curve, has long been a foundational principle in machine learning \citep{6797087}. However, the emergence of the ``double descent'' \citep{nakkiran2019deepdoubledescentbigger} phenomenon has challenged this view, demonstrating that test error can decrease as model complexity increases into the highly overparameterized regime \citep{Belkin_2019, nakkiran2019deepdoubledescentbigger}. 

In this work, we also conduct a fine-grained analysis of this behavior specifically in modern text classification, exploring how different model architectures AR, MLM, DIFF, and ENC with varying data and model sizes. The results depicted in Figure~\ref{fig:iterate_over_datasets} illustrate that the classical bias-variance trade-off is predominantly observed in small-data settings (i.e., lower sample sizes), reinforcing the foundational principle in these regimes. Our findings offer critical empirical insights that nuance the prevailing theory. While we did not scale model size sufficiently to observe the full double descent curve, our results are consistent with the overparameterization regime. Specifically, as shown in Figure~\ref{fig:iterate_over_datasets}, the classical \textbf{``inverted U''} phenomenon is only evident in small-data settings. In contrast, for full-data settings, model performance is consistently \textbf{\textit{non-decreasing}} with increasing model size, suggesting that larger models do not necessarily perform poorly when trained on ample data. This trend is particularly pronounced for the AR architecture, which exhibits a robustly non-decreasing performance curve across our experiments.

Moreover, Figure~\ref{fig:iterate_over_strategy} visualizes the interplay between sample size and strategy more explicitly. We further investigate the \textbf{``more data can hurt''} phenomenon \citep{nakkiran2019deepdoubledescentbigger}, finding that while more data generally improves performance, a performance drop is occasionally observed in specific datasets such as \texttt{sst5} and \texttt{multiclasssentiment}. This finding, illustrated in Figure~\ref{fig:iterate_over_strategy}, suggests that although the phenomenon exists, no universal dataset-specific pattern dictates its occurrence.

\section{Validation-loss checkpointing is canonical in discriminative classification}
\label{app:val_loss_canonical}

Selecting checkpoints by validation loss/error is a standard early-stopping protocol in supervised learning. Prechelt’s \citep{prechelt1997earlystop,prechelt2012earlystop} early stopping description (\S1.2) states: “Use the weights the network had in that previous step as the result of the training run.” 
Bengio’s training recommendations state that the selected iteration “should be the point of lowest validation error in the training run.” \cite{DBLP:journals/corr/abs-1206-5533}
Modern empirical studies of deep learning training practice explicitly define best-checkpoint selection by validation loss; e.g., \cite{PhamQWLRTYN20}\ define “Best-loss selection criterion” as selecting the checkpoint with the “lowest validation loss.” 
\cite{lu2021clam}\ similarly state that “the saved model, which has the lowest validation loss, is then tested on the test set.” 
\cite{pmlr-v162-mindermann22a}\ note: “We always use the IL model checkpoint with lowest validation loss (not highest accuracy) …” 
\cite{ingle-etal-2022-investigating}\ write: “We utilize validation loss as the metric to choose the best checkpoint.”

\section{Acknowledgments}
The research and writing of this paper were led by Siva Rajesh Kasa, who oversaw the entire process from conception and experimental design through to implementation strategies, core experiments, analytical insights, and final manuscript approval. Karan Gupta managed end-to-end diffusion model experiments, owned the rebuttal and camera-ready cycles, and played a significant role in dataset details, implementation, preparing bias-variance analysis, and guiding the integration of calibration and ordinality sections. Sumegh Roychowdhury took the lead on noise robustness studies, compiled and presented the main results, generated key tables and figures, and coordinated the overall paper structure to ensure clarity and narrative cohesion across all sections. Ashutosh Kumar was instrumental in designing and implementing both the MLM and robustness experiments, contributed to pre-trained and encoder model evaluations, and co-authored substantial parts of the introduction, methodology, and related work sections, as well as leading dataset-centric analyses and diffusion studies. Yaswanth Biruduraju was responsible for all aspects of the calibration and ordinality experiments, including their conception, implementation, and analysis, ensuring these metrics were rigorously evaluated and reported. Santhosh Kumar Kasa built the core experimental infrastructure, particularly the discrete diffusion model implementation, ran experiments for both generative and discriminative paradigms, and maintained comprehensive codebase documentation to guarantee reproducibility. Nikhil Priyatam Pattisapu contributed the autoregressive and pseudo-generative methodologies, led the associated empirical studies on model and sample size scaling, authored relevant sections, and played a constructive role in team consensus around generative versus discriminative distinctions and multiple manuscript revisions. Arindam Bhattacharya anchored the theoretical framing by shaping the generative-discriminative comparison paradigm and was essential in literature review and conceptual foundations. Shailendra Agarwal and Vijay Huddar supported discussions, and provided manuscript feedbacks

\end{document}